\documentclass{article}

\PassOptionsToPackage{numbers, compress}{natbib}

\usepackage[preprint]{neurips_2026}

\usepackage[T1]{fontenc}
\usepackage{graphicx}
\usepackage{comment}
\usepackage{amsmath,amssymb} %
\usepackage{color}
\usepackage{xcolor}
\usepackage{booktabs}
\usepackage{subcaption}
\usepackage[colorlinks=true]{hyperref}
\usepackage{bm}
\usepackage{xspace}
\usepackage{dsfont}
\usepackage{graphics}

\usepackage{textcomp}
\usepackage{microtype}

\usepackage[all]{nowidow}

\usepackage{algorithm,algorithmic}
\usepackage{wrapfig,epsfig}
\usepackage{mathtools}
\usepackage{dsfont}
\usepackage{graphicx}
\usepackage{nomencl}
\usepackage{multirow}
\usepackage{makecell}
\usepackage{adjustbox}
\usepackage{subcaption}
\usepackage{enumerate}
\usepackage{enumitem}
\usepackage{multirow}
\usepackage{amssymb}
\usepackage[symbol]{footmisc}
\usepackage{colortbl}
\usepackage{diagbox}
\usepackage{fontawesome}

\usepackage{booktabs}
\usepackage{bbding}
\usepackage{pifont}
\usepackage{wasysym}
\usepackage{utfsym}

\usepackage{xcolor}

\usepackage{hyperref}
\definecolor{modernCite}{RGB}{20, 50, 200}
\definecolor{modernLink}{RGB}{140, 0, 210}  
\definecolor{modernURL}{RGB}{220, 100, 0}   

\hypersetup{
    colorlinks=true,         
    citecolor=modernCite,   
    linkcolor=modernLink,   
    urlcolor=modernURL,   
    pdftitle={My Document},  
}

\usepackage{tcolorbox}
\tcbuselibrary{minted,skins}
\usepackage{float}

\definecolor{codegreen}{rgb}{0.0, 0.5, 0.4}
\definecolor{codegray}{rgb}{0.5, 0.5, 0.5}
\definecolor{codepurple}{rgb}{0.58, 0, 0.82}
\definecolor{codeblue}{rgb}{0.0, 0.4, 0.7}
\definecolor{codered}{rgb}{0.7, 0.1, 0.1}
\definecolor{backcolour}{rgb}{1.0, 1.0, 1.0}

\usepackage{listings}
\lstdefinestyle{pythonstyle}{
    backgroundcolor=\color{backcolour},
    commentstyle=\color{codegreen},
    keywordstyle=\color{codeblue}\bfseries,
    stringstyle=\color{codered},
    basicstyle=\ttfamily\small,
    breakatwhitespace=false,
    breaklines=true,
    captionpos=t,
    keepspaces=true,
    showspaces=false,
    showstringspaces=false,
    showtabs=false,
    tabsize=4,
    frame=none,
    xleftmargin=2mm,
    xrightmargin=2mm,
    aboveskip=0pt,
    belowskip=0pt,
    emph={self, True, False, None},
    emphstyle=\color{codeblue},
    morekeywords={def, class, return, raise, for, in, if, else, import, from, as, with},
}
\usepackage[utf8]{inputenc} 
\usepackage[T1]{fontenc}    
\usepackage{hyperref}       
\usepackage{url}            
\usepackage{booktabs}       
\usepackage{amsfonts}       
\usepackage{nicefrac}       
\usepackage{microtype}      
\usepackage{xcolor}         

\title{Differentiable Ray Tracing with Gaussians for Unified Radio Propagation Simulation and View Synthesis}

\author{%
  Niklas Vaara\textsuperscript{1}, Lam Huynh\textsuperscript{2}, Pekka Sangi\textsuperscript{1},  Miguel Bordallo L\'opez\textsuperscript{1}, Janne Heikkil\"a\textsuperscript{1}\\
  \textsuperscript{1} Center for Machine Vision and Signal Analysis, University of Oulu, Finland\\
  \textsuperscript{2} CubiCasa Oy, Finland\\
  \url{https://github.com/nvaara/GaussianRT-RF-NVS}
}

\begin{document}

\maketitle

\begin{abstract}

Explicit neural representations such as 3D Gaussian Splatting (3DGS) enable high-fidelity and real-time novel view synthesis, yet optimize for alpha-composited optical appearance rather than ray-intersectable geometry. In contrast, radio-frequency (RF) digital twins require deterministic multi-bounce paths, where the geometry dictates trajectories and their associated attenuation and delay. We introduce a framework enabling differentiable RF propagation simulation directly within visually reconstructed neural scenes, allowing point-to-point path computation between arbitrary 3D locations while preserving high-quality visual rendering. Unlike conventional RF simulation pipelines that rely on manually constructed meshes, we embed Gaussian primitives into a hardware-accelerated ray tracing structure as the underlying spatial representation. By extracting physically meaningful channel impulse responses from visual-only reconstructions, we provide cross-modal evidence that neural reconstructions can serve as unified spatial representations for both electromagnetic propagation simulation and photorealistic view synthesis.

\end{abstract}

\section{Introduction}
\label{sec:intro}

Cameras and radio transceivers provide complementary observations of the same physical scene: one through visible-light image formation, the other through radio wave propagation. Although these measurements appear very different, they are governed by the same underlying geometry and material interfaces: the surfaces that create image structure also block, reflect, and attenuate radio waves. This makes novel-view synthesis and radio-frequency (RF) simulation two forward models over a shared latent 3D environment.

Recent progress in neural scene representations has made this shared-view increasingly plausible. In particular, 3D Gaussian Splatting~\cite{kerbl20233d} has emerged as a highly efficient explicit representation for reconstructing scenes from images or videos, enabling high-fidelity, real-time novel-view synthesis without manual 3D modeling. Subsequent extensions~\cite{huang20242d, chen2024pgsr, turkulainen2025dn} further improve geometric fidelity through depth supervision, normal regularization, and surface-aware constraints. However, despite their visual quality, standard 3DGS pipelines are designed around rasterization and alpha-composited optical appearance. They do not directly expose the ray-intersectable geometry required to compute physical multi-bounce propagation paths.

This limitation is especially consequential for wireless digital twins. Accurate radio propagation simulation requires deterministic line-of-sight (LoS) and non-line-of-sight (NLoS) multipath trajectories, since these paths determine the channel characteristics used to construct channel frequency responses (CFRs) and channel impulse responses (CIRs). Existing ray tracing (RT) pipelines therefore rely on explicit geometric models, typically CAD or mesh reconstructions, that are expensive to build, difficult to update, and often lack the fine geometric details present in real environments. As a result, the geometric modeling bottleneck remains a major obstacle to scalable, site-specific RF simulation.

The second obstacle is material uncertainty. Radio propagation modeling depends not only on geometry but also on material parameters such as permittivity and conductivity. In practice, these parameters are often incomplete, unavailable, or specified only for limited material classes and frequency ranges, as in standards such as ITU-R P.2040~\cite{itu}. This mismatch can lead to substantial discrepancies between simulated and measured channels. Recent differentiable RT methods~\cite{hoydis2024learning, jiang2025learnable} address this issue by learning material properties from RF measurements, but their effectiveness still heavily depends on having an accurate geometric model of the environment.

These parallel advances reveal a critical problem: computer vision can now reconstruct rich 3D scenes automatically, while RF simulation can increasingly learn electromagnetic (EM) parameters from measurements, but the two remain disconnected by incompatible scene representations. Visual neural reconstructions provide appearance-optimized Gaussian primitives, whereas EM simulation requires deterministic multi-bounce paths over physically meaningful geometry. To bridge this divide, we introduce a differentiable RT framework built directly on Gaussian scene representations. Our method enables multi-bounce point-to-point path computation within visually reconstructed neural scenes, while preserving photorealistic novel-view synthesis and supporting differentiable EM computations that enables learning from RF measurements.

Our contributions are threefold: (1) a unified, differentiable Gaussian representation that jointly supports photorealistic rendering, multi-bounce geometric path tracing, and differentiable electromagnetic modeling; (2) a real-world multimodal dataset comprising RGB-D images, calibrated camera poses, and channel measurements; and (3) a differentiable alignment objective that optimizes channel parameters by matching simulated and measured CIRs. Together, these elements establish a single spatial representation that generalizes across optical and RF domains, enabling scalable, measurement-driven digital twins without manual geometric modeling.

\section{Radio Propagation Preliminaries}
\label{sec:preliminary}

In wireless channels, the received signal is typically composed of multiple propagation paths. These paths from transmitter (TX) to receiver (RX) are due to interactions with the environment via different mechanisms such as reflection. These paths fulfill Fermat's principle of least time, and can effectively be modeled with rays. Each path, in addition to the time delay $\tau_i$, is modeled with a complex path coefficient $a_{i}$ computed, similarly to \cite{hoydis2024learning}, using
\begin{equation}
    a_{i} = \frac{\lambda}{4\pi}\mathbf{G}_{rx}(\theta_{rx,i}, \phi_{rx,i})^{\mathsf{H}} \mathbf{T}_{i} \mathbf{G}_{tx}(\theta_{tx,i},\phi_{tx,i}),
\end{equation}
where $\lambda$ is the wavelength 
and $\mathbf{G}_{tx}$ and $\mathbf{G}_{rx}$ are the antenna patterns of TX and RX, respectively.
Both patterns are functions of the azimuth $\phi$ and elevation $\theta$ angles, with output in $\mathbb{C}^{2 \times 1}$. The symbol $\mathbf{T}_{i}$ denotes the transfer matrix that contains the effect of all EM interactions of the $i$th path, including free-space path loss. In the case of a specular reflection, the transfer matrix of that $k$th interaction is defined as \cite{aoudia2025sionna}
\begin{equation*}
    \mathbf{T}_{k}^r =
    \begin{bmatrix}
        r_{\perp} & 0 \\
        0 & r_{\parallel}
    \end{bmatrix}
    \begin{bmatrix}
        \mathbf{e}_{pe}^\top \mathbf{k}_{u} & \mathbf{e}_{pe}^\top \mathbf{k}_{v} \\
        \mathbf{e}_{pa}^\top \mathbf{k}_{u} & \mathbf{e}_{pa}^\top \mathbf{k}_{v}
    \end{bmatrix}, \qquad
        \mathbf{e}_{pe} = \frac{\mathbf{k} \times \mathbf{n}}{||\mathbf{k} \times \mathbf{n} ||_ 2}, \qquad
    \mathbf{e}_{pa} = \mathbf{e}_{pe} \times \mathbf{k},
\end{equation*}
where $r_{\perp}$ and $r_{\parallel}$ are the perpendicular and parallel reflection coefficients, $\mathbf{k}_{u}$ and $\mathbf{k}_{v}$ are arbitrary unit direction vectors orthonormal to the incident ray direction $\mathbf{k}$, and $\mathbf{n}$ is the surface normal. For an incident wave in vacuum, the corresponding Fresnel reflection coefficients are
\begin{equation*}
    r_{\perp}
    = \frac{\cos\theta_{i}
        - \sqrt{\epsilon_{c} - \sin^{2}\theta_{i}}}
           {\cos\theta_{i}
        + \sqrt{\epsilon_{c} - \sin^{2}\theta_{i}}}, \qquad
    r_{\parallel}
    = \frac{\epsilon_{c} \cos\theta_{i}
        - \sqrt{\epsilon_{c} - \sin^{2}\theta_{i}}}
           {\epsilon_{c} \cos\theta_{i}
        + \sqrt{\epsilon_{c} - \sin^{2}\theta_{i}}}.
\end{equation*}

Here, $\theta_{i}$ is the angle of incidence, and $\epsilon_{c}$ is the complex relative permittivity of the reflecting material, expressed as
$
    \epsilon_{c} = \epsilon_{r}
    - j\,\frac{\sigma_{c}}{\omega \epsilon_{0}},
$
where $\epsilon_{r}$ is the real relative permittivity, $\sigma_{c}$ is the conductivity, and $\epsilon_{0}$ is the vacuum permittivity.

Based on these definitions, the channel frequency response (CFR) $H$ at frequency $f$ is obtained with
\begin{equation}
\label{eq:fft}
    H(f) = \sum_{i=1}^N a_{i} e^{-j2\pi f \tau_{i}}\text{,}
\end{equation}
where $N$ is the number of paths. In practice, Eq.~\ref{eq:fft} is sampled at uniformly spaced frequencies, forming a discrete spectrum. The equivalent CIR in the time-domain can be obtained via the inverse discrete Fourier transform. From this CIR, the power delay profile (PDP) is formed by taking the squared magnitude of each CIR sample.

\section{Related Work}

\subsection{Differentiable Scene Representation}

Differentiable scene representations have shifted from implicit neural fields such as NeRF~\cite{mildenhall2021nerf} to explicit point-based methods. 3D Gaussian Splatting (3DGS)~\cite{kerbl20233d} replaces costly volumetric ray marching with an optimized tile-based rasterizer, enabling real-time photorealistic view synthesis. However, as standard 3DGS projects semi-transparent ellipsoids onto the image plane, it lacks the geometric structure needed to model physical transport phenomena such as multi-bounce reflections.

To address the limited physical interaction of probabilistic volumes, several works extract surface topology or enable ray intersections within the Gaussian framework. SuGaR~\cite{guedon2024sugar} aligns Gaussians with underlying surfaces to recover continuous meshes, while 2D Gaussian Splatting~\cite{huang20242d} flattens primitives into surface-aligned disks for improved depth and multi-view consistency.

Parallel efforts introduce physically based rendering into explicit Gaussian representations. GaussianShader~\cite{jiang2024gaussianshader} and Relightable 3D Gaussians~\cite{gao2024relightable} decouple materials from lighting, whereas IRGS~\cite{gu2024irgs} and SpecTRe-GS~\cite{tang2025spectre} replace rasterization with BVH-based RT to support reflections, shadows, and dynamic relighting. Hardware-accelerated RT approaches such as 3DGRT~\cite{moenne20243d} and EnvGS~\cite{xie2025envgs} further improve efficiency and support complex reflective effects. While these methods enable visible-light rendering, their application to non-visible electromagnetic wave propagation remains largely unexplored. Building on explicit ray–Gaussian intersection, we adapt differentiable Gaussian representations to support the simulation of multi-bounce radio propagation.

\subsection{Differentiable Radio Frequency Methods}

Neural and non-neural scene representations have been explored for RF modeling, with NeRF \cite{mildenhall2021nerf} inspiring several channel modeling approaches. WineRT \cite{orekondy2023winert} predicts ray-surface interactions using a NeRF-like surrogate but relies on synthetic triangle meshes. NeRF-2 \cite{zhao2023nerf2} combines RT with an MLP to learn spatial RF distributions, while NeWRF \cite{haofan2024newrf} uses direction-of-arrival cues and ray search for channel prediction with fewer measurements.

The success of 3DGS \cite{kerbl20233d} has motivated explicit Gaussian-based RF field modeling. WRF-GS \cite{wen2025wrf} employs spherical and Mercator projections for differentiable channel synthesis, and WRF-GS+ \cite{wen2025neural} extends this with deformable Gaussians for static and dynamic components. Complex-valued Gaussians were introduced in \cite{yang2025gsrf} for faster training and inference. RF-3DGS \cite{zhang2026rf} adopts a two-stage optimization of radiance, geometry, and radio fields, and integrates real measurements with ray-traced simulations from synthetic meshes.

Implicit methods rely heavily on measurements and synthetic data and are expensive to train, while 3DGS-based approaches offer improved efficiency. However, except for WineRT \cite{orekondy2023winert}, these methods cannot explicitly recover geometry-dependent propagation paths. Our method enables both novel-view synthesis and direct computation of geometric RF paths with differentiable electromagnetic modeling from an explicit radiance field.

\subsection{Ray Tracing-Based Radio Propagation}

Detailed handcrafted models are costly to produce, motivating the use of reconstructed 3D models for RT-based radio propagation. Triangle meshes reconstructed from point clouds are widely used \cite{okamura2020simplification,pang2021gpu,niu20223d,suga2023indoor,kamari2023environment,xia2024path,suga2025rgb}, typically approximating large planar surfaces to remain compatible with ray tracers such as Sionna \cite{hoydis2023sionna} or Wireless Insite \cite{wirelessinsite}.

Laser scanned point clouds offer higher geometric fidelity \cite{virk2015simulating,jarvelainen2016indoor,koivumaki2022point,koivumaki2023ray} but require heavy downsampling and still require multi-minute simulation times. NimbusRT v0 \cite{vaara2025ray} accelerates computation via voxelization and conical rays, while NimbusRT v1 \cite{vaara2025differentiable} models intersections as circular disks with distance-based attenuation and uses a visibility matrix for efficient higher-order reflections. Both rely on surface labels to prune duplicate paths and require hand-tuned primitive sizes and weights.

To enhance the accuracy of radio propagation modeling, Hoydis \textit{et al.} \cite{hoydis2024learning} introduced a differentiable RT framework to calibrate the EM material properties. They use an MLP with positional encoding for improved material modeling, but it removes explicit material parameters, relies on synthetic meshes, and fails to generalize to sparse measurements.
Expanding on this paradigm, \cite{bakirtzis2025radio} explored the scalability of such mesh-based differentiable pipelines for large radio environments. However, these meshes lack important geometric details relevant at higher frequencies. 
NimbusRT v1 \cite{vaara2025differentiable} addressed this by supporting differentiable radio propagation with high fidelity point clouds. 
Despite this, its environment model remains non‑differentiable, leaving the method sensitive to sensor noise and dependent on hand‑tuned parameters. In contrast, our method provides a fully differentiable scene representation and uses semantic segmentation information to boost robustness under sparse channel measurement data. In addition, it supports photorealistic rendering capabilities absent in prior RT-based radio propagation simulators.

\section{Method for Differentiable Gaussian Ray Tracing}

\label{sec:gs_rt}
Our method utilizes a set of RGB, semantic, normal, and depth images, accompanied by an optional confidence map, as well as the camera poses, and a sparse point cloud. The objective is to reconstruct a geometrically accurate 3D scene representation that supports both photorealistic novel view synthesis and physically consistent multi-bounce RT simulations. To achieve this, the point cloud is represented using 2D Gaussians, which are turned into triangle meshes for RT. We optimize them with a differentiable hardware accelerated ray tracer together with photometric and geometric regularizations. A high level illustration is provided in Fig.~\ref{fig:overview} and implementation details can be found in Appendix~\ref{appendix:rt_details}.

\subsection{Gaussian Representation}

Our method follows 2D Gaussian Splatting (2DGS)~\cite{huang20242d}, which models splats as planar Gaussians with zero extent along the $z$-axis. This makes them well suited for RT, since they behave as elliptical disks and allow precise ray-Gaussian intersections. Each Gaussian is defined by parameters $[\mathbf{\mu}, \mathbf{s}, \mathbf{q}, \mathbf{h}, \sigma, \gamma]$: the mean position $\mathbf{\mu}$, scale $\mathbf{s}=[s_x,s_y]$, rotation quaternion $\mathbf{q}$, spherical harmonics $\mathbf{h}$ for view-dependent color, opacity~$\sigma$, and edge probability~$\gamma$. We additionally assign each Gaussian a material label~$\ell$ as a post-processing step.

\subsection{Ray Tracing}

\begin{figure*}
    \centering
    \includegraphics[width=1.0\linewidth]{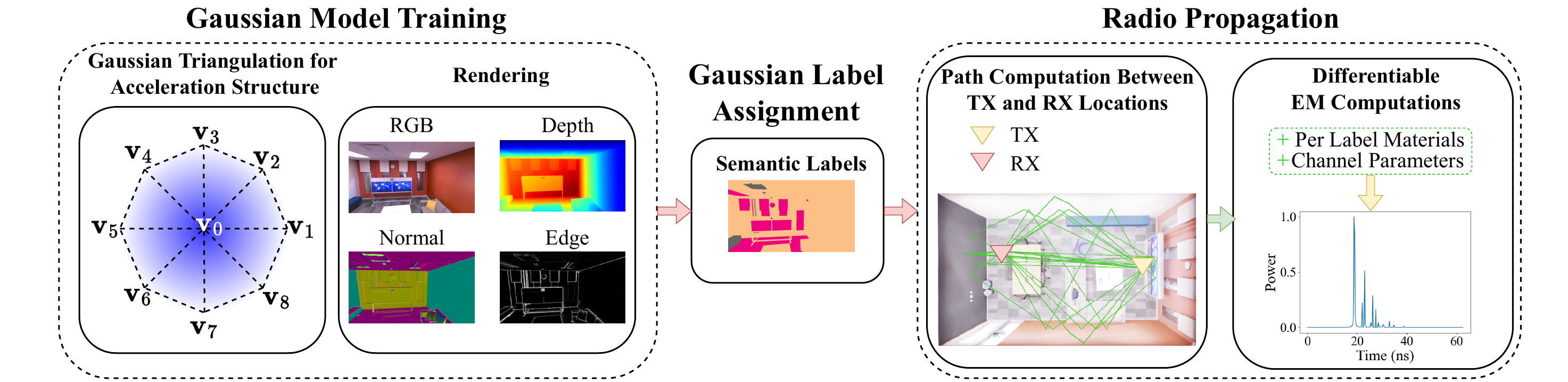}
    \caption{Our method represents 2D Gaussians using a triangle‑based representation and renders RGB, normal, depth and edge images. After optimization, semantic labels are assigned to each Gaussian. The annotated model is then used to resolve propagation paths via RT, and these paths subsequently serve as input for differentiable electromagnetic computations.}
    \label{fig:overview}
\end{figure*}
We first find the ray-plane intersection $\mathbf{i}$, which is followed by evaluating the Gaussian local space 2D intersection coordinates $(u ,v)$:
\begin{equation*}
    \mathcal{G} = \exp\left(-\frac{u^2 + v^2}{2}\right), \qquad
    u = \frac{\mathbf{r}_{x}^{\top} (\mathbf{i} - \mu)}{s_{x}}, \quad
    v = \frac{\mathbf{r}_{y}^{\top} (\mathbf{i} - \mu)}{s_{y}} \text{,}
\end{equation*}
where $\mathbf{r}_{x}$ and $\mathbf{r}_{y}$ are the first and second column of the rotation matrix derived from $\mathbf{q}$.
We perform ray-Gaussian intersections using a hardware‑accelerated ray tracer that alpha‑blends Gaussians in the exact depth order. To enable efficient intersection testing, we approximate each Gaussian with a lightweight polygonal mesh and sort intersected Gaussians in small batches of Gaussians until ray saturation, as in \cite{moenne20243d}. We cap each ray to 128 intersections and cache them to avoid a second RT pass during backpropagation. Full details of our Gaussian construction and intersection handling are provided in Appendix~\ref{appendix:rt}.

\paragraph{Alpha Blending.}

Alpha blending-based rendering is performed as in standard 3DGS \cite{kerbl20233d}. The $i$th Gaussian is evaluated for its alpha contribution $\omega_{i}$ in front-to-back order as follows:
\begin{equation*}
    \omega_i = T_i \sigma_i \mathcal{G}_i, \qquad
    T_i = \prod_{j=1} ^ {i-1} (1 - \alpha_{j})\text{.}
\end{equation*}
where $T_{i}$ is the transmittance. Finally, the alpha blended color $\mathbf{c}$ of a ray intersecting with $N$ Gaussians is then computed as 
$    \bar{\mathbf{c}} = \sum_{i} \omega_{i} \mathbf{c}_{i}$,
where $\mathbf{c}_{i}$ is the view-dependent color of the $i$th Gaussian obtained from $\mathbf{h}$. Similarly, the normal maps $\bar{\mathbf{n}}$ are computed with
\begin{equation}
    \label{eq:wsa}
    \bar{\mathbf{n}} = \frac{\sum_{i} \omega_{i} \mathbf{n}_{i}}{\max\left(\epsilon, \sum_{i}\omega_{i}\right)}
\end{equation}
where $\mathbf{n}$ denotes the Gaussian surface normal, which is the third column of the rotation matrix derived from $\mathbf{q}$, and $\epsilon$ is a small constant to avoid division by zero. For each ray, the intersection distances $t_{i}$ and edge probabilities $\gamma_{i}$ of each Gaussian are aggregated using the same weighted-sum averaging scheme as in Eq.~\ref{eq:wsa}, giving the distance map $\bar{t}$ and edge map $\bar{\gamma}$.

\paragraph{Ray Label Determination.}
\label{sec:label}
As our application is RT, we prioritize a labeling strategy that is both explicit and computationally lightweight. We retrieve the strongest contributor among the $N$ intersected Gaussians by selecting the one with the highest per-ray weight $\omega_{i}$. This corresponds to choosing the Gaussian that contributes most to the ray under the standard 3DGS alpha blending. For semantic label generation, we use Dinov3~\cite{simeoni2025dinov3} head trained on ADE20K \cite{zhou2017scene} dataset. For details on how the labels are assigned, please see Appendix~\ref{appendix:label}.

\subsection{Training}

The training starts from a sparse point cloud, obtained from RGB images and known camera poses. We initialize these sparse points as Gaussians, which are optimized using gradient descent. During training, our method produces alpha blended colors, depths, and normals, which are utilized in the loss functions for visual and geometric consistency. We build on prior densification and pruning strategies with minor modifications. Full implementation details are provided in Appendix~\ref{appendix:densify_prune}.

\paragraph{Normal Regularization.}
We implement normal consistency to ensure that the Gaussians are aligned with the actual surface. By assuming local planarity, pseudo normals $\mathbf{n}^{d}$ can be computed from the rendered depth map gradients using finite differences \cite{huang20242d}. These depth map normals guide the Gaussian normal directions to match the underlying surface, resulting in smooth surfaces, as demonstrated in \cite{huang20242d, yu2024gaussian}. To further improve the results especially near edges, gradients computed from the rgb image may be used to weight the regularization \cite{chen2024pgsr, turkulainen2025dn}. Thus, following these works, we employ a normal consistency loss  
\begin{equation}
    \mathcal{L}_{n} = \frac{1}{N_{p}} \sum_i \mathcal{I}_{i}(1 - |\bar{\mathbf{n}}_{i}^\top \mathbf{n}_{i}^{d}|),
\end{equation}
where $N_{p}$ is the number of pixels and $\mathcal{I}$ is the edge weight computed from the normalized RGB image gradient as in \cite{chen2024pgsr}:
\begin{equation}
    \mathcal{I}_{i} = (1 - \nabla c_{rgb})^2
\end{equation}
However, such regularization alone results in artifacts in highly reflective areas, as noted in \cite{xie2025envgs, turkulainen2025dn}. Similar to the aforementioned works, we utilize monocular normal estimates $\mathbf{n}^{m}$ computed using MoGe-2 \cite{wang2025moge} to supervise the rendered normals maps with
\begin{equation}
    \mathcal{L}_{m} = \frac{1}{N_{p}}\sum_i \mathcal{I}_{i}(1 - |\bar{\mathbf{n}}_{i}^\top \mathbf{n}_{i}^{m}|)\text{.}
\end{equation}

\paragraph{Depth Regularization.}
Standard 3DGS~\cite{kerbl20233d} learns Gaussian parameters using only photometric supervision, which is insufficient for accurate geometry, especially in textureless or reflective regions. We add depth regularization on ray traced depth maps, following Turkulainen \textit{et al.}~\cite{turkulainen2025dn}, who show that logarithmic loss yields smoother reconstructions. When depth confidence maps are available, we further restrict the loss to pixels with maximum confidence. This gives the following loss:
\begin{equation}
    \mathcal{L}_{d} = \frac{1}{N_{p}} \sum_{i} \mathcal{I}_{i} \mathcal{B}_{i}\log\left(1 + ||\bar{d}_{i} - d_{i}||_1\right),
\end{equation}
where $\mathcal{B}_{i}$ is the binary mask derived from the confidence map, $d_{i}$ is the ground truth depth for the $i$th pixel, and $\bar{d}_{i}$ is the estimated depth obtained from the Euclidean distance map $\bar{t}_{i}$.

\paragraph{Edge Regularization.}

To obtain reliable geometric edges for RT, we derive a binary edge mask from the monocular normal estimate. We compute the gradient magnitude of the predicted normals and mark a pixel as an edge whenever it exceeds the heuristic threshold $t_{\text{edge}}$. This gives us the binary mask $\mathcal{B}_{\gamma}$, which we use in our edge regularization loss:
\begin{equation}
    \mathcal{L}_{e} = \frac{1}{N_{p}} \sum_i(\mathcal{B}_{\gamma} - \bar{\gamma}_{i})^{2}.
\end{equation}

\paragraph{Total Loss.}

We adopt the RGB reconstruction loss proposed in 3DGS \cite{kerbl20233d}, which combines RGB L1 loss with a D-SSIM term to better capture perceptual differences and structural consistency in the rendered images:
\begin{equation}
    \mathcal{L}_{rgb} = (1 - \lambda_{rgb}) \mathcal{L}_{1} + \lambda_{rgb} \mathcal{L}_{D-SSIM}\text{.}
\end{equation}
The combined total loss is
\begin{equation}
    \mathcal{L}_{gs} = \mathcal{L}_{rgb} + \lambda_{n} \mathcal{L}_{n} + \lambda_{m} \mathcal{L}_{m} + \lambda_{d} \mathcal{L}_{d} + \lambda_{e} \mathcal{L}_{e} \text{.}
\end{equation}

\section{Method for Propagation Path Computation with Gaussians}
\label{sec:prop}
The input to our method consists of the labeled Gaussian model pipeline presented in Section~\ref{sec:gs_rt}, along with the Tx and Rx positions. The objective is to identify LoS and multi-bounce propagation paths that fulfill the Fermat's principle of least time. 

\subsection{Path Computation}

Our method is inspired by the RT-based based radio propagation simulation tool NimbusRT \cite{vaara2025differentiable}, which operates on top of point clouds. The path computation begins by identifying coarse propagation paths. These coarse paths are then refined to satisfy Fermat’s principle of least time. Finally, the refined paths are post‑processed to remove duplicates and ensure a clean set of propagation paths.

\paragraph{Coarse Path Computation.}

For efficient coarse path computation, we compute a visibility matrix between each RX and voxel, similar to \cite{vaara2025differentiable}. This allows us to simply query whether an RX is visible from the intersected volume instead of an expensive RT operation. We utilize the gaussian mean points to form a voxel grid with a given resolution $v_{r}$. We refer to these voxels containing at least one Gaussian mean as discretized elements (DEs).

Once the visibility matrix is computed, we then process each TX separately. For each TX, we cast a ray to each DE, which keeps the number of rays small compared to casting potentially millions of rays. Each ray that intersects with a surface is then reflected specularly based on the intersection normal $\bar{\mathbf{n}}$. After each interaction, we then query the visibilty matrix to see whether the RX is visible from that location. In case it is, we save the path as a candidate for path refinement.

\paragraph{Path Refinement.}

The coarse paths are refined by minimizing their total length following \cite{vaara2025ray, vaara2025differentiable}, assuming locally planar interactions and optimizing each interaction point in its tangent plane. The full parameterization and update rules are provided in Appendix~\ref{appendix:path_refinement}.

\paragraph{Path Pruning.}

When multiple refined paths converge to nearly identical interaction points, they represent the same physical specular reflection. Thus, keeping all of them would overcount its contribution. While mesh‑based methods can prune duplicates via triangle indices \cite{aoudia2025sionna}, point cloud methods face noisy geometry, making existing distance‑ or hashing‑based strategies \cite{jarvelainen2016indoor, koivumaki2022point, vaara2025ray, vaara2025differentiable} either slow or error prone. We instead exploit the sparsity of the environment‑driven RT and identify unique paths directly from geometry. For $N$ candidates, all $N(N-1)/2$ pairs are tested: two paths are considered duplicates if each interaction point lies within distance $g_{d}$ of the other point’s plane defined by its normal, and if their normals differ by less than $g_{n}$. Additionally, we prune paths with interactions near edges by exploiting the edge information $\gamma$ embedded in each Gaussian. Implementation details can be found in Appendix~\ref{appendix:edge_aware}.

\section{Experiments}
\label{sec:experiments}
Our method is implemented using custom CUDA‑based RT kernels and Slang-Torch \cite{slangtorch}. In this section, we focus on evaluating the radio propagation simulations. The experiments were ran on a PC equipped with an NVIDIA GeForce RTX 5090 GPU. All of the parameters are provided in Appendix~\ref{appendix:hyperparameters}. In Appendix~\ref{appendix:additional_experiments}, we provide additional experiments, and show that our method achieves similar quality in novel view synthesis compared to existing Gaussian-based methods.

\subsection{Validation with NLoS D-Band Channel Measurements}
\begin{wrapfigure}[12]{r}{0.35\textwidth}
    \centering
    \includegraphics[width=\linewidth]{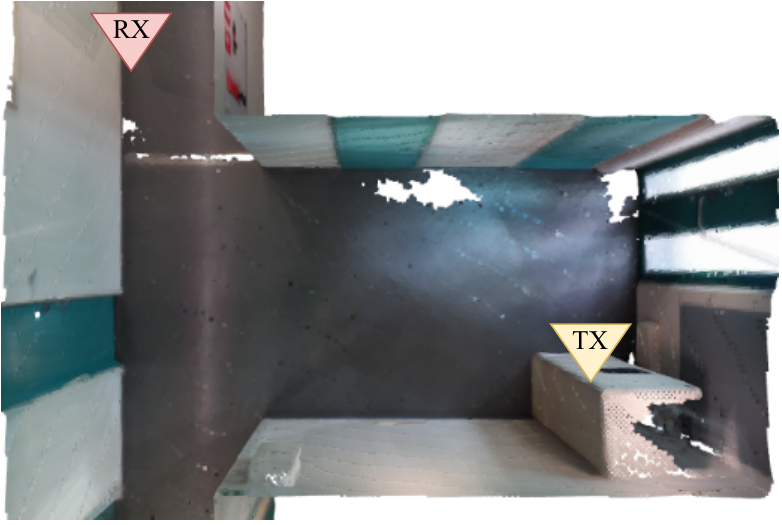}
    \caption{Corridor scene with NLOS measurements.}
\end{wrapfigure}
\paragraph{Measurements and Environment.}
We compare against the channel measurements conducted in \cite{kokkoniemi2022initial}. The measurements consist of a NLoS case, where the frequency was swept in 4001 samples from 110~GHz to 170~GHz in many directional measurements. The geometry was captured with a RGB-D camera. We follow the setup provided in \cite{vaara2025ray}, where the measurements are composed into an aggregated CIR by choosing the strongest sample from all of the directional measurements, and all materials are assumed to be plasterboard. More details and an ablation study are provided in Appendix~\ref{appendix:corridor_scene}.

\paragraph{Baselines and Evaluation Metrics.}
The primary comparison class for our method is pipelines that utilize reconstructed 3D models. Thus, we compare against NimbusRT v0 \cite{vaara2025ray}, and its successor NimbusRT v1 \cite{vaara2025differentiable}, which are point cloud-based ray tracers for radio propagation simulation. Handcrafted mesh simulators such as Sionna~\cite{hoydis2023sionna} and Wireless Insite~\cite{wirelessinsite} are included as reference points under manually prepared geometry, rather than as baselines.
As our evaluation metric, we utilize RMS delay spread $\tau_{rms}$, defined as
\begin{equation*}
\tau_{\mathrm{rms}} = \sqrt{\frac{\sum_{i} p_i (\tau_i - \bar{\tau})^2}{\sum_{i} p_i}}, \qquad \bar{\tau} = \frac{\sum_{i} p_i \tau_i}{\sum_{i} p_i},
\end{equation*}
where $p_{i} = |h_{i}|^2$ and $\tau_{i}$ are the power and delay of $i$th tap in the CIR, respectively, and $\bar{\tau}$ is the mean excess delay.

\paragraph{Results.}

\begin{figure}[]
    \centering

    \begin{subfigure}{0.32\linewidth}
        \centering
        \includegraphics[width=\linewidth]{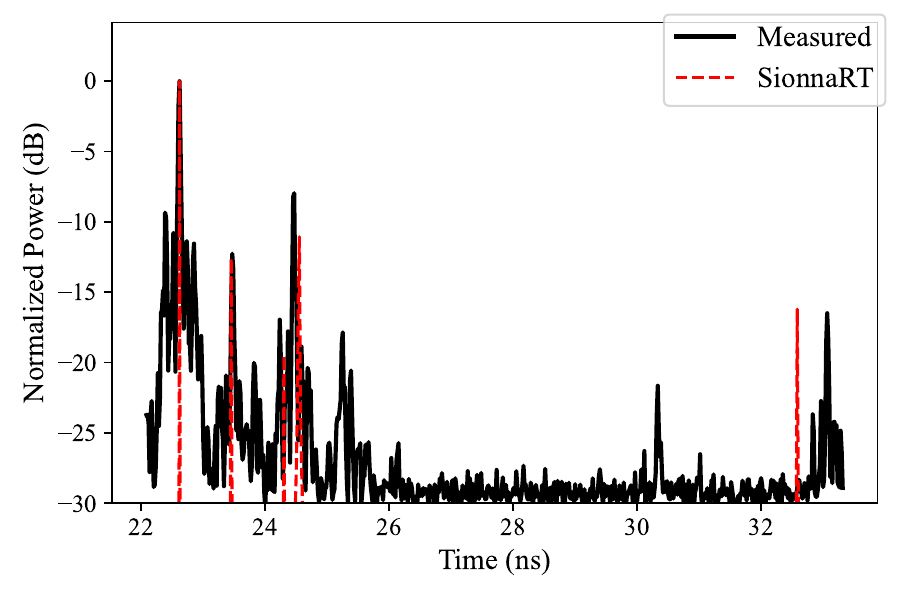}
        \caption{Sionna}
        \label{fig:sionna_pdp}
    \end{subfigure}
    \hfill
    \begin{subfigure}{0.32\linewidth}
        \centering
        \includegraphics[width=\linewidth]{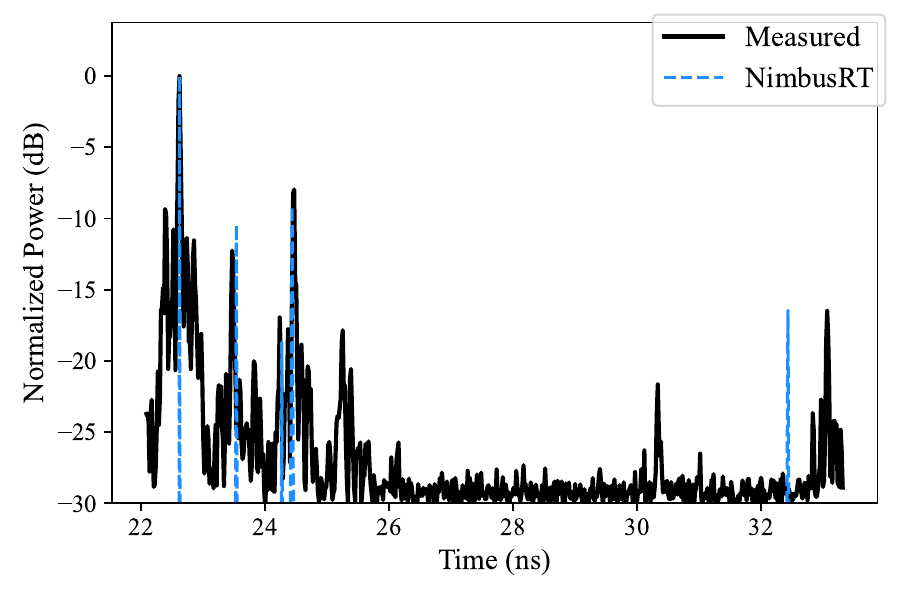}
        \caption{NimbusRT v1}
        \label{fig:nimbus_v1_pdp}
    \end{subfigure}
    \hfill
    \begin{subfigure}{0.32\linewidth}
        \centering
        \includegraphics[width=\linewidth]{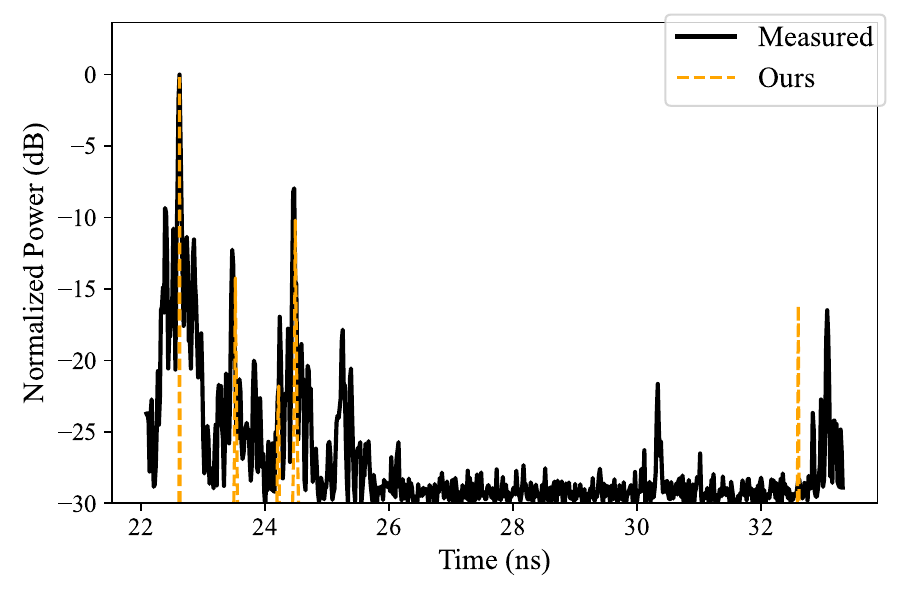}
        \caption{Ours}
        \label{fig:ours_pdp}
    \end{subfigure}

    \caption{Max normalized PDPs of NimbusRT v1, Sionna, and our method, compared with the measured one.}
    \label{fig:PDPs}
\end{figure}

We provide the qualitative results of simulated PDPs of Nimbus v1, Sionna and our method overlayed on top of the measured PDPs in Fig.~\ref{fig:PDPs}. All three methods produce similar peaks as in the measured one. We further evaluate the RMS delay spread of each method in Table~\ref{tab:rms_ds}. Our method is capable of producing more similar $\tau_{rms}$ than NimbusRT in comparison to the simulations with synthetic triangle mesh model and channel measurements. This result highlights the effectiveness of a differentiable environment representation augmented with geometric regularization.

\subsection{Experiments with Differentiable Radio Propagation}

Our objective is to maintain a fully explicit and interpretable representation of the propagation process. We therefore optimize geometric and material parameters solely to reproduce the measured PDPs, acknowledging that many different combinations of relative permittivity and conductivity can yield similar amplitude responses and that extracting precise material properties is often difficult and time consuming. For complete details of these experiments, please refer to Appendix~\ref{appendix:hall_scene}.
\begin{wrapfigure}[12]{r}{0.3\textwidth}
    \centering
    \includegraphics[width=\linewidth]{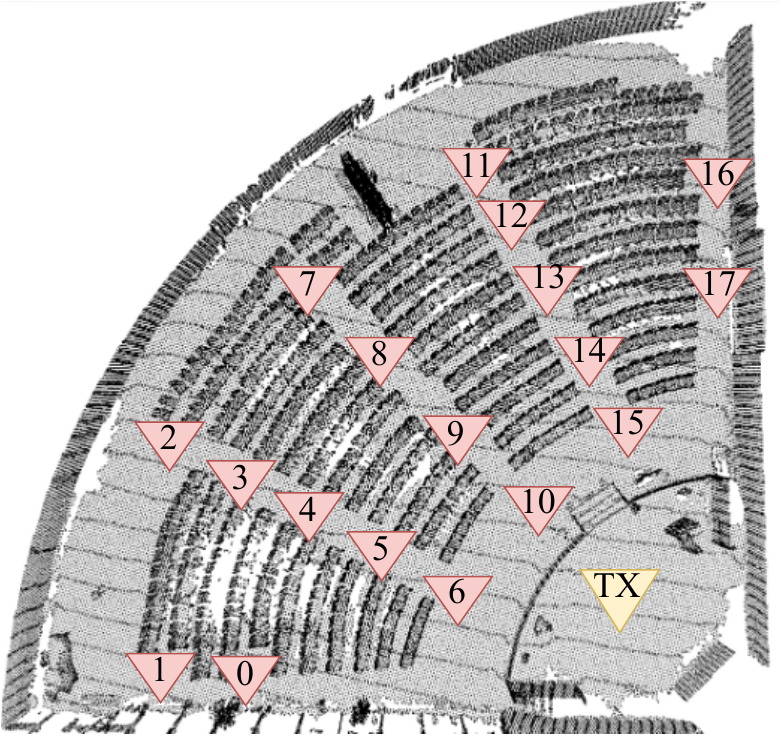}
    \caption{Auditorium scene with one TX and 18 RXs.}
    \label{fig:hall_scene}
\end{wrapfigure}

\paragraph{Measurements and Environment.}

The channel measurements were carried out in a large auditorium at 234~GHz center frequency. The 4~GHz bandwidth surrounding the center frequency was swept in 1001 uniformly spaced frequency samples. The setup consisted of one TX position and 18 RX positions, as illustrated in Fig.~\ref{fig:hall_scene}. The geometry was captured using a RGB-D camera.

\begin{table}[]
    \centering
    \caption{Results of simulated $\tau_{rms}$. The results marked with * were taken from \cite{vaara2025ray}.}
    \resizebox{1.0\textwidth}{!}{
    \begin{tabular}{lcccccc}
        \toprule
         Metrics & Measurements* & Nimbus v0* & Wireless Insite* & Nimbus v1 & Sionna & Ours \\
         \midrule
         $\tau_{rms}$ (ns) & 1.67 & 1.52 & 1.58 & 1.52 & 1.57 & 1.56\\
         Absolute Error (ns) & - & 0.15 & 0.09 & 0.15 & 0.10 & 0.11\\
         \bottomrule
    \end{tabular}
    }
    \label{tab:rms_ds}
\end{table}

\paragraph{Structural Path Alignment Loss.}

\begin{wrapfigure}[25]{r}{0.4\textwidth}
    \centering
    \includegraphics[width=\linewidth]{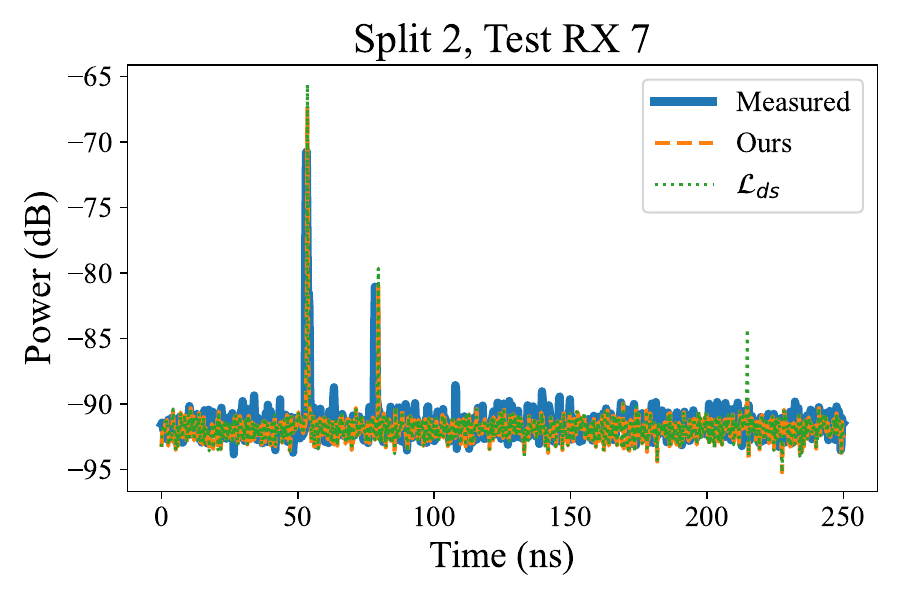}
    \caption{Unlike $\mathcal{L}_{ds}$, our loss reduces the contribution of the late arriving path to the noise level, which is not observable in the measurements.}
    \label{fig:success_pdp}
    \includegraphics[width=\linewidth]{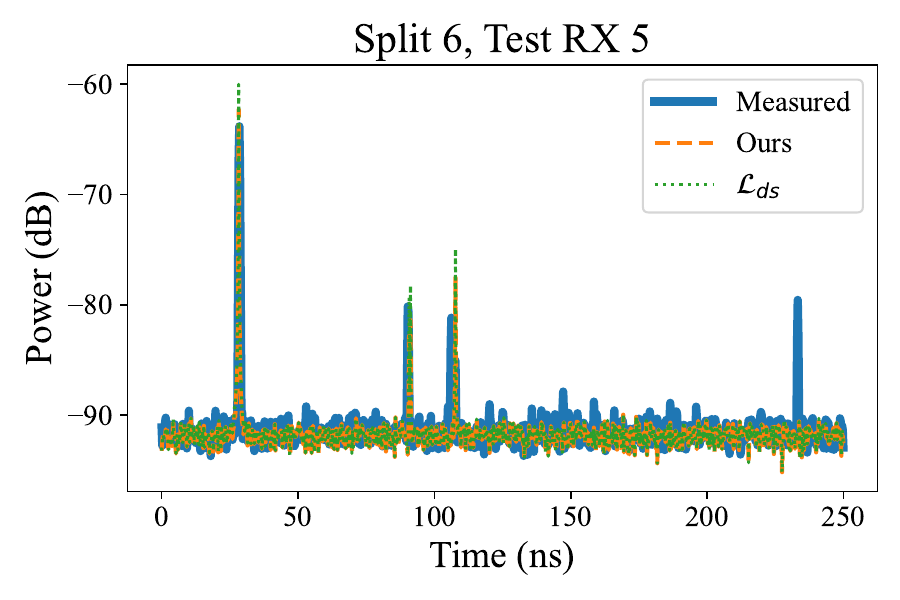}
    \caption{Our method misses a late arriving propagation path, leading to a substantial underestimation of $\tau_{rms}$.}
    \label{fig:failure_pdp}
\end{wrapfigure}

We begin by extracting LoS and reflection multipath components (MPCs) from noise by thresholding the measurements at $-88$~dB (see Fig.~\ref{fig:success_pdp}). Because adjacent taps above threshold often form clusters, we prune each cluster by retaining only its strongest MPC. We then compute the signed distance (SD) to the nearest MPC to enable reliable correspondence matching between measured and simulated paths, preventing the optimizer from locking onto weaker scatterers. A maximum SD threshold is applied to further avoid erroneous MPC associations. We formulate our loss as
\begin{equation}
    \mathcal{L}_{spa} = \frac{1}{N}\sum_{i=1}^N \frac{| p_{i} - G_{ch} \hat{p}_{i} |}{ p_{i} + G_{ch} \hat{p}_{i}},
\end{equation}
where $N$ is the number of paths, $G_{ch}$ is the channel gain and $p_{i}$ and $\hat{p}_{i}$ are the ground truth and predicted powers of the $i$th correspondence, respectively. The power values are extracted from the measured and simulated CIRs based on the time delay of each simulated path.

\paragraph{Training.}

We train the model for 1000 iterations using the Adam optimizer with a learning rate of $1\text{e-}{2}$. At each training iteration, we randomly pick one RX from the training set. We optimize the relative permittivity $\epsilon_r$ and conductivity $\sigma_c$, and the channel gain coefficient $G_{\text{ch}}$. To ensure that all material parameters remain physically valid, we optimize unconstrained variables $\epsilon_r'$, $\sigma_{c}'$, and $G_{\text{ch}}'$ and map them to valid values via
\begin{equation*}
    \epsilon_r = 1 + \exp(\epsilon_{r}'), \qquad
    \sigma_c = \exp(\sigma_{c}'), \qquad
    G_{ch} = \exp(G_{ch}').
\end{equation*}
For evaluation, we use the 18 RX locations and construct six distinct train-test splits. In each split, three RX locations are held out for evaluation while the remaining fifteen are used for training. The test RX indices follow a cyclic pattern: the first split uses RX locations $[0, 6, 12]$, the second uses $[1, 7, 13]$, and so on. In the simulation, we assume the antenna pattern to be an isotropic antenna pattern at both TX and RX side, and restrict the simulation to LoS and a maximum of two reflections.

As a baseline, we also train using the RMS delay spread–based loss introduced in \cite{hoydis2024learning}.
\begin{equation}
    \mathcal{L}_{ds} = \frac{|\tau_{rms}- \hat{\tau}_{rms}|}{\tau_{rms} + \hat{\tau}_{rms}} + \frac{|\mathcal{P} - \hat{\mathcal{P}} | }{\mathcal{P} + \hat{\mathcal{P}} },
\end{equation}
where $\mathcal{P}$ and $\hat{\mathcal{P}} = G_{ch} \sum_i p_i$ are ground truth and predicted total power, respectively.
\paragraph{Results.}

The quantitative results are provided in Table~\ref{tab:rms_ds_testset}. The baseline loss $\mathcal{L}_{ds}$ \cite{hoydis2024learning} aligns closely with the evaluation metric. Thus, it can give the false impression of improved accuracy even when the underlying channel response is not well reproduced. In contrast, our loss attempts to find matching paths, providing a more faithful and physically meaningful supervision, especially if the objective is to reproduce similar reflections as observed in measurements, as shown in Fig.~\ref{fig:success_pdp}.
Nevertheless, our loss $\mathcal{L}_{spa}$ on average achieves a better MAE of $\tau_{rms}$ with the test sets than the baseline. In certain cases, our method fails to find propagation paths observed in the channel measurements. An example of this can be seen in Fig.~\ref{fig:failure_pdp}, leading to an absolute $\tau_{rms}$ error of 17.56~ns. Additional results for the test cases and ablation study of $\mathcal{L}_{spa}$ are provided in Appendix~\ref{appendix:hall_scene}.

\subsection{Discussion}
\paragraph{Limitations.}
Our simulations do not consider diffraction, scattering, penetration, and temporal effects. Some paths may be missed due to the coarse RT required for duplicate path removal. Experimental validation is limited to two real‑world scenes, as to the best of our knowledge, no datasets provide RGB‑D images paired with channel measurements.

\paragraph{Broader Impact.}
This work integrates Gaussian-based scene reconstruction with differentiable physically grounded RF point-to-point path tracing, a step towards RF digital twins. These capabilities can benefit wireless communications, robotics, and sensing through multipath characterization, and can also be used to generate synthetic datasets for training neural networks.

\begin{table}[H]
    \centering
    \caption{MAE of $\tau_{rms}$ per test set in nanoseconds.}
    \begin{small}
    \begin{tabular}{lccccccl}
        \toprule
         Method & Set 1 & Set 2 & Set 3 & Set 4 & Set 5 & Set 6 & Mean \\
         \midrule
         Initial & 11.80 & 6.90 & 5.51 & 7.54 & 10.62 & 13.12 & 9.25 \\
         With $\mathcal{L}_{ds} \cite{hoydis2024learning}$ & \bf{7.38} & 6.03 & 3.27 & 6.82 & \bf{6.70} & \bf{9.02} & 6.54 \\
         With $\mathcal{L}_{spa}$ (Ours) & 7.97 & \bf{4.70} & \bf{3.20} & \bf{5.7}2 & 6.85 & 9.66 & \bf{6.35}\\ 
         \bottomrule
    \end{tabular}
    \end{small}
    \label{tab:rms_ds_testset}
\end{table}

\section{Conclusion}

We introduced a unified Gaussian representation that integrates hardware-accelerated ray tracing with an explicit Gaussian scene representation, enabling accurate multi-bounce RF propagation path computation directly from visually reconstructed environments. By bridging optical reconstruction and wireless channel modeling, our approach removes the need for manual CAD geometry and supports differentiable optimization of parameters involved in radio propagation. Our results demonstrate that physically meaningful CIRs can be recovered from a photorealistic scene representation, opening opportunities for tighter integration between vision-based scene understanding and RF simulation.

\paragraph{Future Work.}
Future work includes incorporating diffraction by leveraging the per‑Gaussian edge probability, adding diffuse scattering and learned antenna patterns. We also aim to stabilize path refinement and expand evaluation to larger and more diverse real‑world environments.

\begin{ack}

We would like to thank Naveeth Lafir, Dr. Peize Zhang, and Dr. Pekka Kyösti for performing, providing, and helping with the processing of the measurement data in the auditorium scene. We also thank Dr. Joonas Kokkoniemi for the measurement data of the corridor scene and Dr. Janne Mustaniemi for capturing and reconstructing the auditorium and corridor scenes. Lastly, we thank Shakthi Gimhana, Taufiq Ahmed, Dr. Praneeth Susarla, and Dr. Dileepa Marasinghe for the setup and measurement data in the laboratory scene. 

This research was supported by the Research Council of Finland (former Academy of Finland) 6G Flagship Programme (Grant Number: 346208), and the Horizon Europe CONVERGE project (Grant 101094831).

\end{ack}

\bibliographystyle{unsrtnat}
\bibliography{main}


\appendix

\section{Details of Gaussian Ray Tracing}
\label{appendix:rt_details}

In this appendix, we detail the components underlying our approach, including the RT details, the procedure for semantic label assignment, and the densification and pruning strategy used during Gaussian optimization.

\subsection{Gaussian Mesh Construction}
\label{appendix:rt}
To enable efficient hardware-accelerated intersection testing, we approximate each 3D Gaussian with a lightweight polygonal mesh. Each Gaussian is represented by its mean $\mu$ and eight uniformly spaced vertices on the unit circle in its local tangent frame. These vertices form a coarse boundary approximation of the Gaussian support region, as illustrated in Fig.~\ref{fig:overview}.

Each vertex $\mathbf{v}_i$ is transformed into world space via, similar to \cite{moenne20243d}:
\[
\mathbf{v}_{i}
= \mu + \mathbf{R}
\left(
\mathbf{D}_{s}\sqrt{\max(0,\,2\log(\sigma))}
\right)\mathbf{b}_{i},
\]

where $\mathbf{b}_i$ denotes the $i$-th unit-circle vertex in local space, $\mathbf{R}$ is the Gaussian rotation matrix, and $\mathbf{D}_{s}$ is the diagonal scale matrix derived from the scale parameters. Following \cite{moenne20243d}, the scaling is adjusted such that the proxy boundary corresponds to the 3DGS visibility threshold of $1/255$. This polygonal proxy allows us to use hardware-accelerated ray–triangle intersection tests while for the most part preserving the spatial extent of each Gaussian.

\paragraph{Intersection Processing and Sorting.}

Our objective is to find the intersection point $t$ with each Gaussian. As the 2D Gaussians are planar, we  determine the intersection point by a ray-plane intersection with the help of the rotation matrix $\mathbf{R} = [\mathbf{r}_{x}, \mathbf{r}_{y}, \mathbf{r}_{z}]$ derived from $\mathbf{Q}$ with the following equation:
\begin{equation*}
    t = \frac{\mathbf{r}_{z}^{\top} (\mathbf{\mu} - \mathbf{r}_o)}{\mathbf{r}_{z}^{\top} \mathbf{r}_d}\text{,}
\end{equation*}
from which the intersection point can be computed with 
\begin{equation*}
    \mathbf{i} = \mathbf{r}_{o} + \mathbf{r}_{d} t,
\end{equation*}
where $\mathbf{r}_{o}$ and $\mathbf{r}_{d}$ are the ray origin and ray direction, respectively.

Our ray tracer processes Gaussian intersections in a manner similar to \cite{moenne20243d}. For each ray, we collect up to 16 intersected Gaussians at a time, sort them by depth, and alpha-blend them in order. This procedure is repeated iteratively until the ray saturates or no further intersections are found.

Unlike prior work, we cap the number of intersections per ray to 128. All intersection results are cached and reused during the backward pass, eliminating the need for a second RT operation and significantly reducing the training time overhead, as shown in Table~\ref{tab:training_time}.

\subsection{Semantic Label Assignment}
\label{appendix:label}
\begin{figure}[H]
    \centering
    \includegraphics[width=1.0\linewidth]{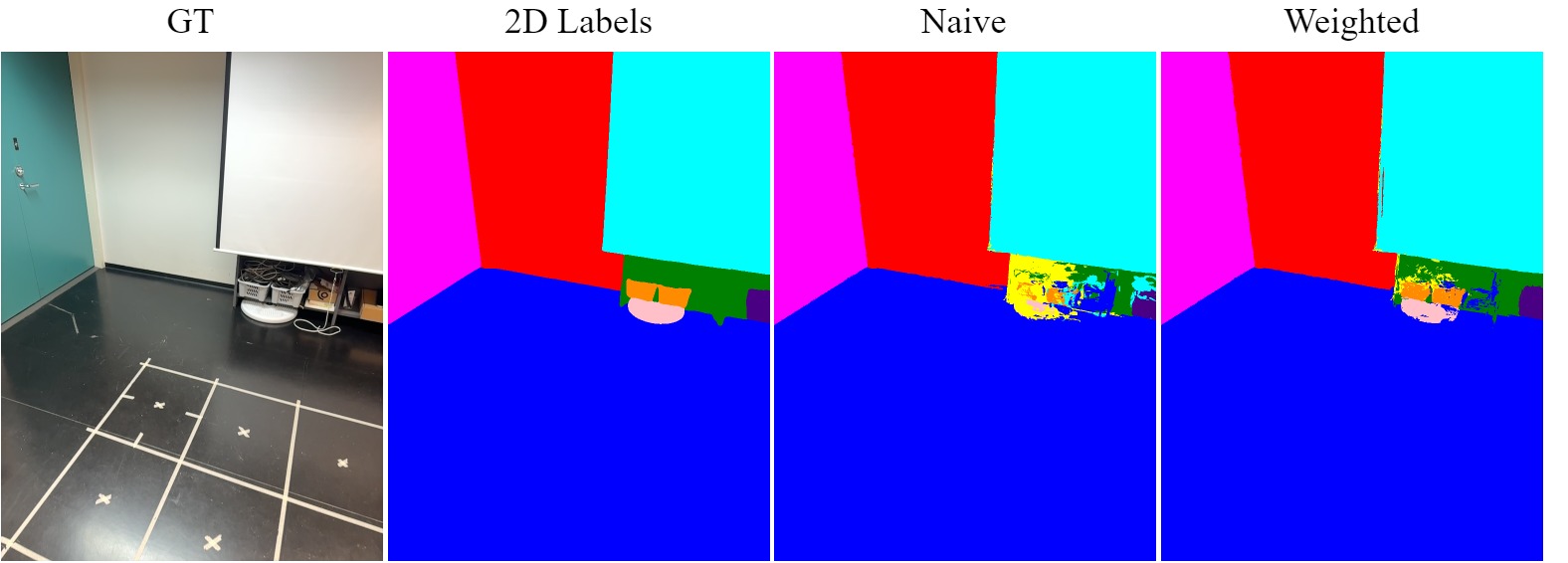}
    \caption{RGB image, 2D semantic labels, and rendered labels with naive and alpha weighted assignment.}
    \label{fig:labeling}
\end{figure}

We use Dinov3~\cite{simeoni2025dinov3} head trained on ADE20K \cite{zhou2017scene} dataset
to generate 2D semantic labels, which are not multi-view consistent. For each view, we compute the per-Gaussian weights $\omega_i$, aggregate them by label, and select the label with the highest total accumulated weight.
We compare against a naive assignment in which each visible Gaussian contributes equally, independent of its weight $\omega_i$.
We provide qualitative results of the labeling strategy in a scene with many objects, which leads to varying per‑view 2D labels. As shown in Fig.~\ref{fig:labeling}, the weight‑based labeling retrieves more accurate per-Gaussian labels. We also exploit this post-processing step to prune all Gaussians that were not seen in any view.

\subsection{Densification and Pruning}
\label{appendix:densify_prune}
\paragraph{Densification.}

As the starting point of the optimization is a spare point cloud, it cannot accurately represent the whole scene. In the original 3DGS, the Gaussians are split and cloned to address the under- and over-reconstruction of the scene \cite{kerbl20233d}. During the densification step, the Gaussians to be densified are picked based on the magnitude of the NDC space positional gradients of the Gaussians.
However, in these methods each Gaussian may contribute to multiple pixels, leading to gradient collision, as noted by \cite{ye2024absgs} and \cite{yu2024gaussian}. The former authors propose computing the absolute value of each gradient component, while the latter authors accumulate the norm of the individual pixel gradients. These approaches demonstrated improved reconstruction quality.
For RT-based Gaussian methods, NDC space gradients do not exist, as the intersections happen in world space. This was addressed in \cite{moenne20243d} by multiplying the Gaussian mean position gradients by half of the distance to the camera. Inspired by these works, we compute take the absolute value of the position gradient, and scale it by half of the distance to the camera.

In \cite{moenne20243d}, the number of primitives are limited to three million. In our implementation, we keep the number of Gaussians uncapped, which leads to our observation of densification instability, especially in outdoor scenes. We observe that this instability is caused by small Gaussians far away from the camera, where their accumulated gradients always pass the densification threshold. To counter this, we clamp the contribution of per-ray densification to $0.99 t_{densify}$. In addition, we address the cases where a Gaussian may degenerate to a line, which often leads to severe over-reconstruction, producing dense Gaussian clusters. Similar behavior was reported in \cite{liu2024citygaussianv2}, which addressed the problem by excluding Gaussians that fulfill $\min(s_x, s_y) / \max(s_x,s_y) \leq t_{s}$, where $t_{s}$ is the axis ratio threshold. We adopt the same filtering strategy.

\paragraph{Pruning.}
We adopt the pruning strategy of 3DGS \cite{kerbl20233d}, where Gaussians with $\sigma < 0.05$ are removed every 500 iterations, as our training ray tracer uses the same tile-based architecture we similarly remove the large screen space Gaussians. In addition to the opacity reset of 3DGS, we also include the opacity decay proposed by Radl \textit{et. al.} in \cite{radl2024stopthepop}, where the opacity of each Gaussian is multiplied by $0.9995$ every 50 iterations. As noted by Radl \textit{et. al.}, this significantly reduces the number of Gaussians.

\section{Details of Path Refinement}
\label{appendix:path_refinement}
To ensure that the initially detected coarse propagation paths are physically consistent and geometrically accurate, we apply a dedicated refinement stage that adjusts each interaction point to better satisfy the Fermat's principle of least time. By refining the interaction points through local optimization, we obtain paths that more faithfully represent the actual propagation behavior and can be reliably used for radio propagation modeling.

The set of coarse paths are refined by minimizing the path length as in \cite{vaara2025ray, vaara2025differentiable}. For this task, local planarity of each point is assumed, where the $k$th interaction point is defined as
\begin{equation}
    \mathbf{I}_{k} = \mu + \mathbf{u}_{k}s_{k} + \mathbf{v}_{k}t_{k},   
\end{equation}
where $\mathbf{u}_{k}$ and $\mathbf{v}_{k}$ are the orthonormal basis vectors of the normalized $\bar{\mathbf{n}}_{k}$ and $s_{k}$ and $t_{k}$ are the unknown parameters that are optimized.
The total length of a path consisting of $N$ interactions can is defined as
\begin{equation}
P = \sum_{k=0}^{N} \| \mathbf{I}_k - \mathbf{I}_{k+1} \|,
\end{equation}
where $\mathbf{I}_{0}$ is the TX and $\mathbf{I}_{N+1}$ is the RX. 
For a path consisting of $N$ interactions, the $k$th interaction point, the local path length from previous to next interaction is  
\begin{equation}
    P_{k} = \| \mathbf{I}_{k-1} - \mathbf{I}_{k}\| + \| \mathbf{I}_{k+1} - \mathbf{I}_{k}\|.
\end{equation} 
We then iteratively minimize $P_{k}$ for $k \in [1, N]$ with respect to $(s_{k}, t_{k})$ using gradient descent, until the maximum number of iterations $\kappa_{i}$ was reached, or $\|\nabla P\|^2 < t_{c}$, where $t_c$ is the convergence threshold. Once a path has converged, its interaction points are validated. For the $k$th interaction, this is done by casting a ray from $\mathbf{I}_{k-1}$ to the optimized position. The path is accepted when two conditions hold: the angular difference between the original and updated surface normals remains below $t_{a}$, and the perpendicular distance from the updated intersection point to the plane defined by the original $\mathbf{I}_{k}$ remains below $t_{d}$. If either condition is not met, the refinement is retried up to $\kappa_{r}$ times.

\section{Details of Experiments}
\label{appendix:experiment_details}
In this appendix, we provide additional details of the experiments conducted in the main paper. The experiments should be interpreted in the context of vision-derived RF simulation: the central question is whether automatically reconstructed geometry can support physically meaningful propagation paths. Manual meshes provide useful reference values, but they do not address the scalable mapping problem targeted by our method.

\subsection{Parameters}
\label{appendix:hyperparameters}
Table~\ref{tab:params} contains all parameters used throughout the paper and appendices, including their descriptions, symbols, and assigned values.
\begin{table}[H]
    \centering
    \caption{Descriptions, symbols and values of the parameters presented in the paper.}
    \begin{small}
    \begin{tabular}{ccc}
        \toprule
         Description & Symbol & Value \\
         \midrule
         Edge magnitude threshold & $t_{\text{edge}}$ & 0.2\\
         Voxel resolution & $v_{r}$ & 0.0625~m \\
         Edge probability threshold & $g_{\gamma}$ & 0.5 \\
         Normal threshold & $g_{n}$ & $5^\circ$ \\
         Distance threshold & $g_{d}$ & 0.3~m \\
         Normal coherency threshold & $g_{i}$ & 0.99 \\
         Axis ratio threshold & $t_{s} $ &  0.02 \\
         Refinement convergence threshold & $t_{c}$ & $1\text{e-}4$ \\
         Refinement angle threshold & $t_{a}$ & $1^\circ$ \\
         Refinement distance threshold & $t_{d}$ & 0.01~m \\
         Maximum refinement iterations & $\kappa_{i}$ & 50 \\
         Maximum refinement retry attempts & $\kappa_{r}$ & 10 \\
         RGB regularization coefficient & $\lambda_{rgb}$ & 0.2 \\
         Monocular normal regularization coefficient & $\lambda_{m}$ & 0.05 \\
         Depth normal regularization coefficient & $\lambda_{n}$ & 0.05 \\
         Edge regularization coefficient & $\lambda_{e}$ & 0.05 \\
         Depth regularization coefficient & $\lambda_{e}$ & 0.2 \\
         \bottomrule
    \end{tabular}
    \end{small}
    \label{tab:params}
\end{table}

\subsection{Corridor Scene}
\label{appendix:corridor_scene}

The channel measurements of this scene were performed in \cite{kokkoniemi2022initial}. The dataset contains a NLOS scenario where the frequency is swept over 4001 samples from 110--170~GHz across many directional measurements. Following \cite{vaara2025ray}, we form an aggregated CIR by selecting the strongest sample across all directional measurements. The measurement setup used horn antennas with half power beamwidths (HPBWs) of $10^\circ$ and $9^\circ$ in azimuth and elevation, covering $90^\circ$ in azimuth and $85^\circ$ in elevation at the RX side. The resulting time-domain CIR has very high resolution, making it suitable for ray tracing validation. Simulations were limited to second-order reflections, and simulated paths were aligned to measurements using the strongest peak.

The environment was captured using Microsoft Azure Kinect camera and the poses were estimated using BS3D \cite{mustaniemi2023bs3d}. The dataset consists of 120 images. For a fair comparison with previous works \cite{vaara2025ray}, we do not perform any semantic segmentation and instead assume the same label for all Gaussians. Similarly, this label represents the electromagnetic material parameters of plasterboard at 60~GHz, as it is not defined at 140~GHz in ITU-R P.2040 \cite{itu}. We utilized MoGe2 \cite{wang2025moge} to generate the monocular normal maps. Instead of the raw sensor depth, we use rendered depths from the reconstructed triangle mesh acquired from BS3D, as the depth pixels visible in the RGB FoV was not able to capture the floors.

\paragraph{Results.}

\begin{figure}[H]
    \centering
    \includegraphics[width=1.0\linewidth]{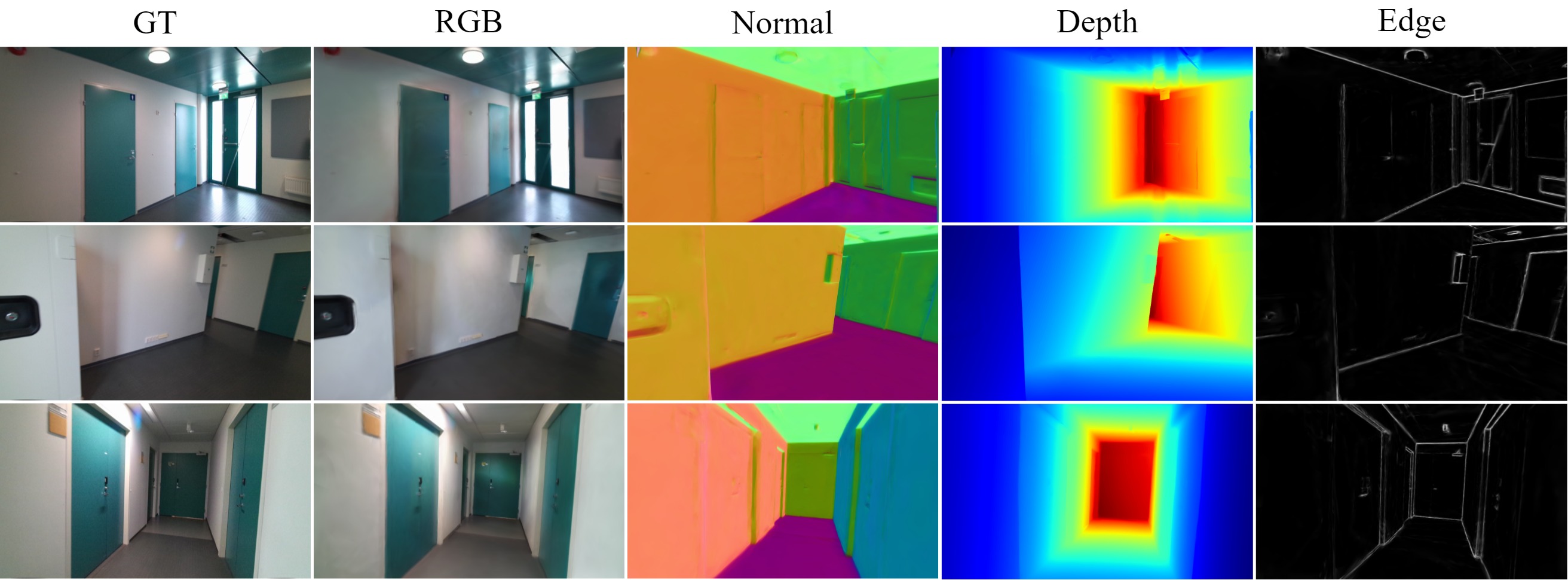}
    \caption{Ground truth RGB image, as well as the rendered RGB, normal, depth and edges by our method in the corridor scene. Zoom in for details.}
    \label{fig:cwc_render}
\end{figure}

After training, the train images had an average PSNR, SSIM, and LPIPS of 24.03, 0.907, and 0.272, respectively. These metric demonstrate mediocre training data quality, however, our method was still capable of producing accurate propagation paths, as shown in Fig.~\ref{fig:ours_pdp}. We provide qualitative rendering of some of the training images in Fig.~\ref{fig:cwc_render}

\paragraph{Ablations.}

\begin{figure}[H]
    \centering

    \begin{subfigure}{0.32\linewidth}
        \centering
        \includegraphics[width=\linewidth]{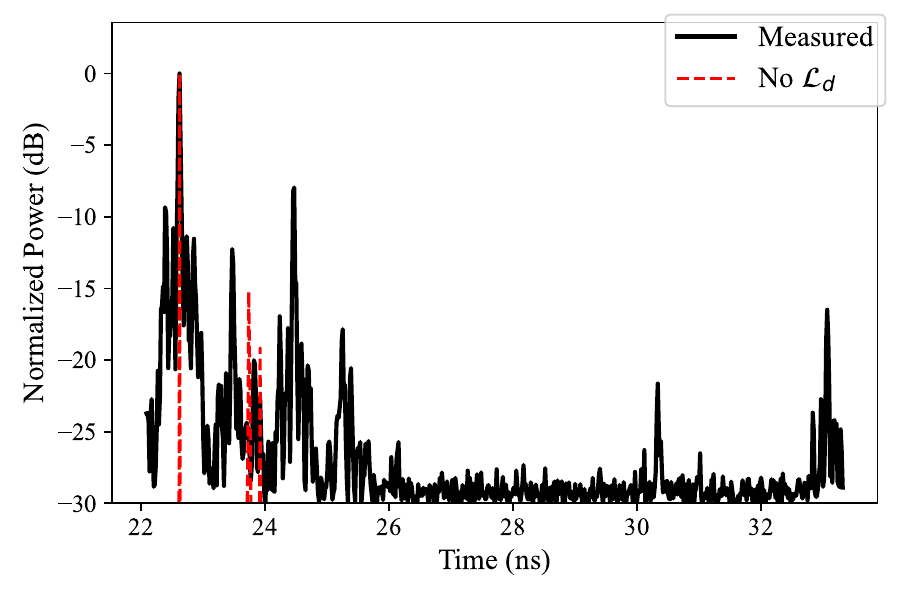}
        \caption{Without $\mathcal{L}_{d}$.}
        \label{fig:wo_ld}
    \end{subfigure}
    \hfill
    \begin{subfigure}{0.32\linewidth}
        \centering
        \includegraphics[width=\linewidth]{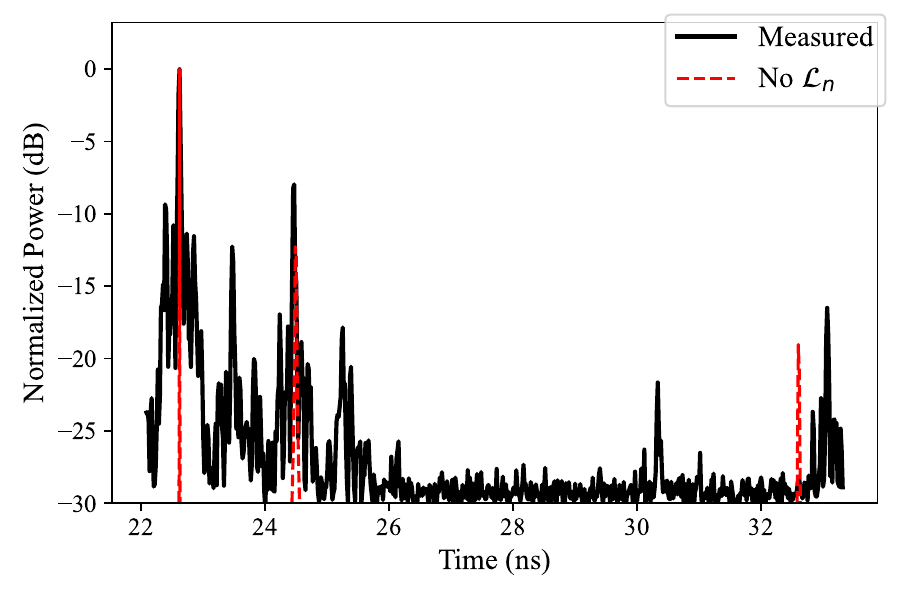}
        \caption{Without $\mathcal{L}_{n}$.}
        \label{fig:wo_ln}
    \end{subfigure}
    \hfill
    \begin{subfigure}{0.32\linewidth}
        \centering
        \includegraphics[width=\linewidth]{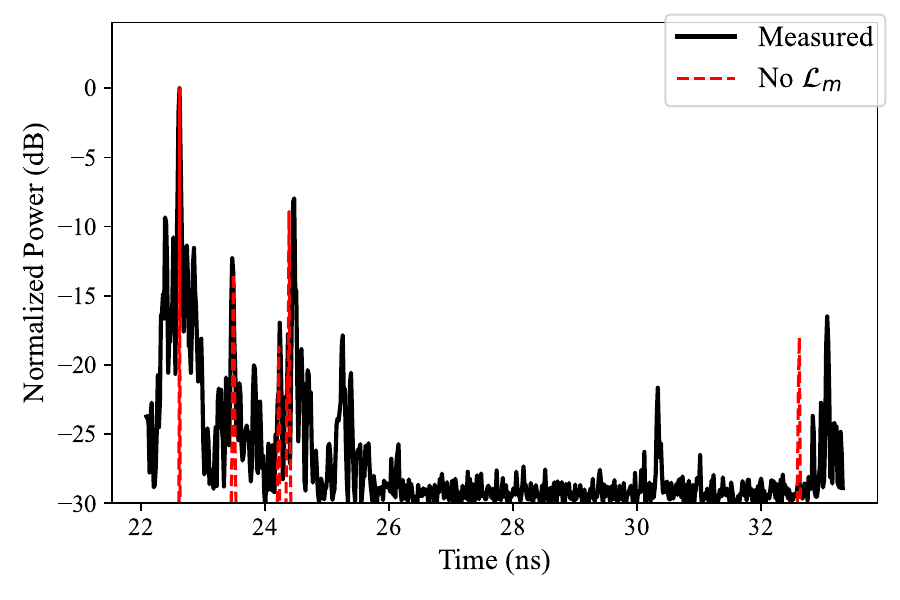}
        \caption{Without $\mathcal{L}_{m}$.}
        \label{fig:wo_lm}
    \end{subfigure}

    \caption{Max normalized power delay profiles of the ablation cases.}
    \label{fig:ablation_pdps}
\end{figure}

We evaluate the effect of normal and depth regularization both in terms of the recovered propagation paths and their impact on $\tau_{\mathrm{rms}}$. Our ablation study isolates the contributions of depth regularization ($\mathcal{L}_{d}$) and the two normal-based terms ($\mathcal{L}_{n}$ and $\mathcal{L}_{m}$). The quantitative results in Table~\ref{tab:ablation_cwc} show that depth regularization is particularly critical for obtaining accurate propagation paths. Although removing $\mathcal{L}_{n}$ results in a $\tau_{rms}$ value numerically closer to the measured one, the qualitative results in Fig.~\ref{fig:ablation_pdps} reveal that several paths are missing, which ultimately increases the simulated $\tau_{rms}$. In contrast, removing $\mathcal{L}_{m}$ produces paths that are visually similar to those of the full model, with only a slight increase in error. Overall, these findings indicate that $\mathcal{L}_{d}$ is essential for accurate geometry reconstruction, and that combining it with depth-based normal smoothing is crucial for reliable path computation.

\begin{table}[H]
    \centering
    \caption{Qualitative results for the ablations in the corridor scene.}
    \begin{small}
    \begin{tabular}{lccccc}
        \toprule
         Metric &  Measured & Without $\mathcal{L}_{d}$ & Without $\mathcal{L}_{n}$ & Without $\mathcal{L}_{m}$ & Full\\
         \midrule
         $\tau_{rms}$ (ns) & 1.67 & 0.21 & 1.58 & 1.54 & 1.56 \\ 
         Absolute Error (ns) & - & 1.46 & 0.09 & 0.13 & 0.11 \\
         \bottomrule
    \end{tabular}
    \end{small}
    \label{tab:ablation_cwc}
\end{table}

\subsection{Auditorium Scene}
\label{appendix:hall_scene}

The channel measurements were conducted in a large auditorium using a VNA-based setup at a 234~GHz center frequency with 4~GHz bandwidth and 1001 uniformly spaced frequency samples. The configuration included one TX position and 18 RX positions (Fig.~\ref{fig:hall_scene}). Both TX and RX employed horn antennas with approximately $19.5^\circ$ HPBW and 19.33\,dBi gain. Azimuth and elevation sweeps were performed in $20^\circ$ increments, except for the RX azimuth, which used a $10^\circ$ step size. On the TX side, the measured azimuth spanned roughly $100^\circ$, ensuring LoS to each RX, while the RX covered the full azimuth range. For elevation, the TX was measured only at $90^\circ$, whereas the RX included two elevations: $70^\circ$ and $90^\circ$.

\begin{figure}[]
    \centering

    \begin{subfigure}{0.32\linewidth}
        \centering
        \includegraphics[width=\linewidth]{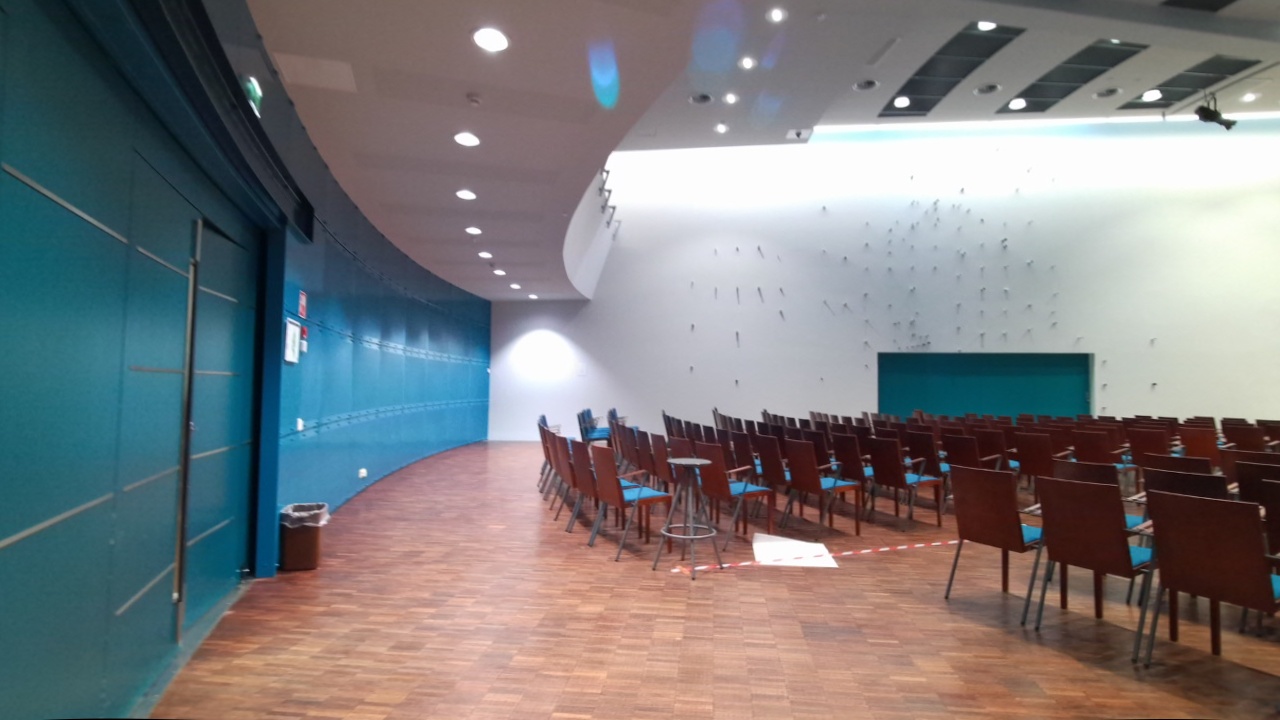}
        \caption{Color image.}
    \end{subfigure}
    \hfill
    \begin{subfigure}{0.32\linewidth}
        \centering
        \includegraphics[width=\linewidth]{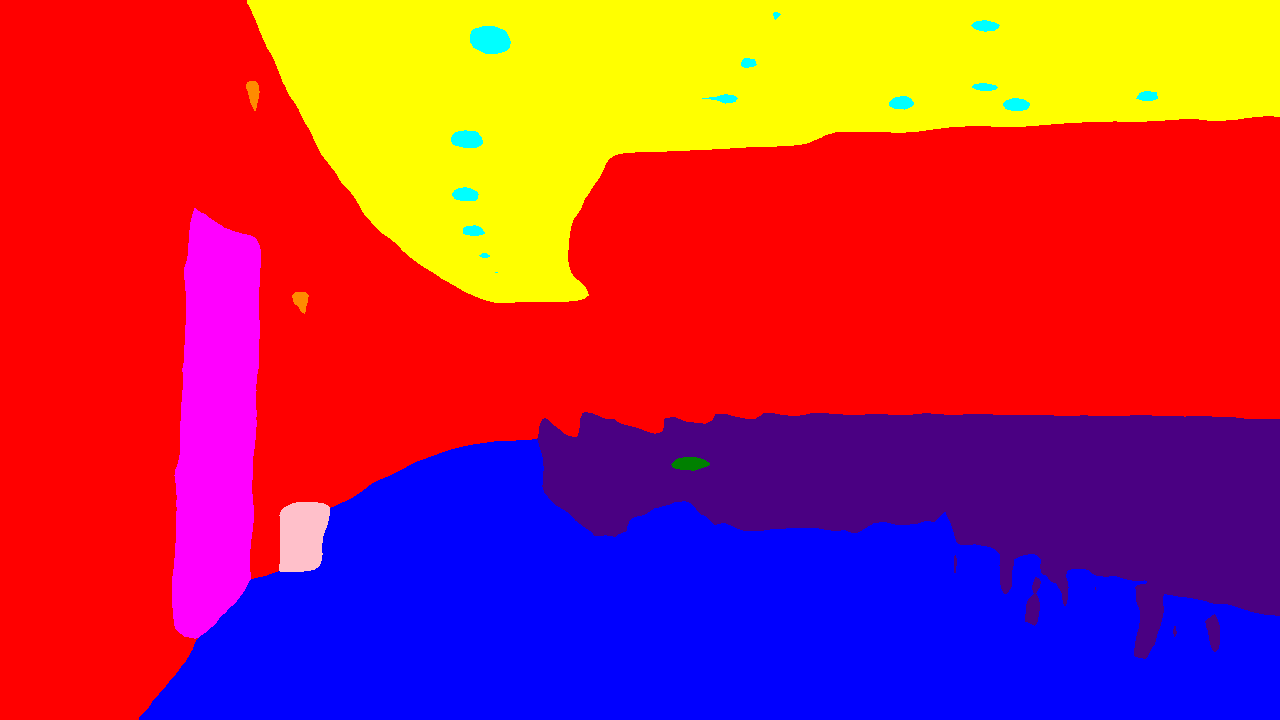}
        \caption{Semantic labels from Dinov3.}
    \end{subfigure}
    \hfill
    \begin{subfigure}{0.32\linewidth}
        \centering
        \includegraphics[width=\linewidth]{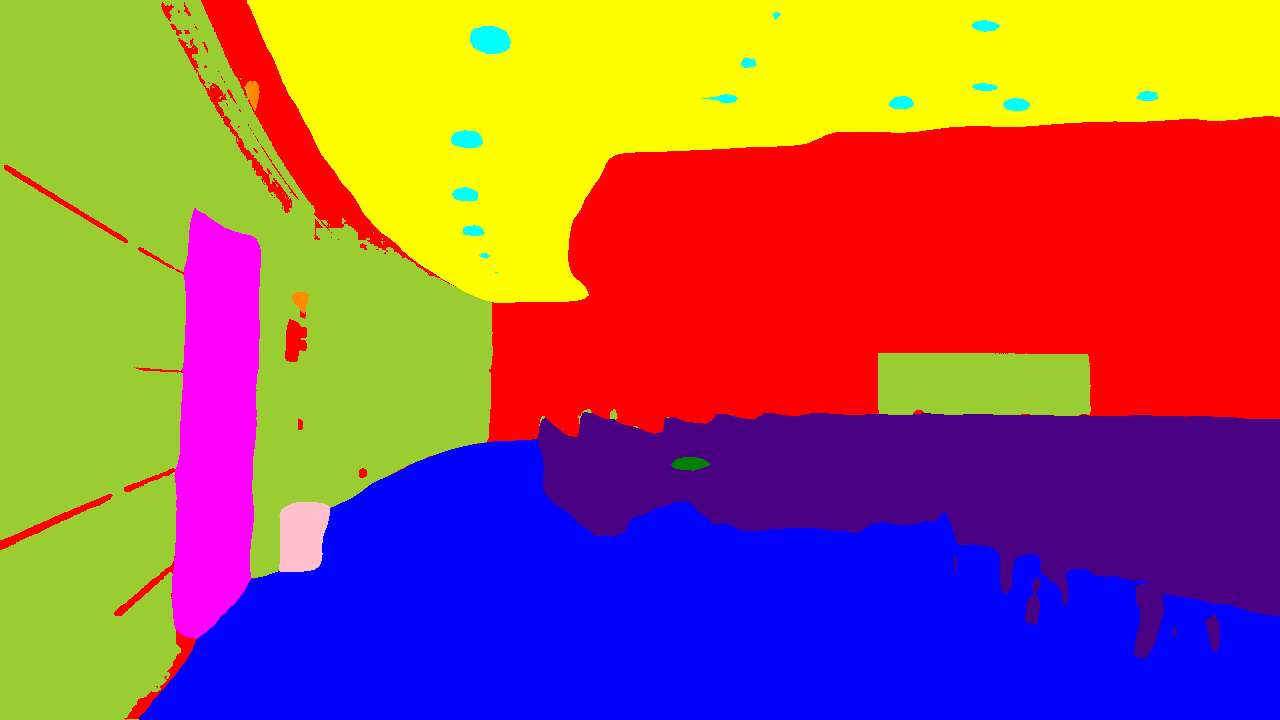}
        \caption{Processed labels.}
    \end{subfigure}

    \caption{Label post-processing.}
    \label{fig:hall_example}
\end{figure}

We utilized Microsoft Azure Kinect to collect RGB-D frames and reconstructed it using BS3D \cite{mustaniemi2023bs3d}. The scene consists of 1314 images which were reduced to 438 RGB, depth and normal images for training. Similar to the corridor scene, we use rendered depths from the reconstructed triangle mesh acquired from BS3D for better depth coverage, and the normals were generated with MoGe2~\cite{wang2025moge}. We also generated 2D segmentation images and assigned them to the Gaussians using the procedure described in Section~\ref{sec:label}. We used Dinov3~\cite{simeoni2025dinov3} head trained on ADE20K \cite{zhou2017scene} dataset to generate 2D segmentation labels. To further improve the labels for the application of material parameter optimization, we processed the predicted wall labels by a simple RGB-based separation to have a separate label for the cyan and white walls (see Fig.~\ref{fig:hall_example}).

\paragraph{Training Details.}
All of the materials were initialized with plasterboard properties computed with the equations provided in ITU-R-P.2040 \cite{itu}, and the gain coefficient $G_{ch}$ was initialized to the peak antenna gain of 19.33~dBi. 

\paragraph{Results.}

\begin{figure}[]
    \centering
    \includegraphics[width=1.0\linewidth]{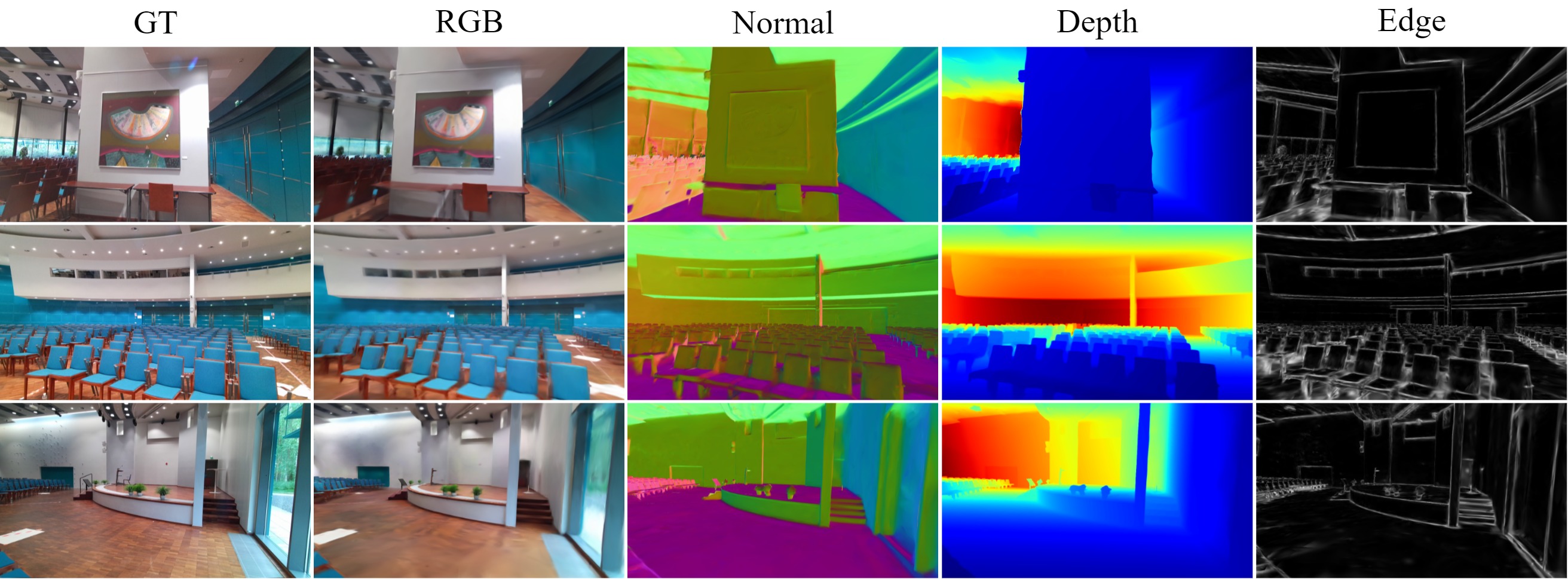}
    \caption{Ground truth RGB image, as well as the rendered RGB, normal, depth and edges by our method in the auditorium scene. Zoom in for details.}
    \label{fig:hall_render}
\end{figure}

\begin{figure}[]
    \centering
    \includegraphics[width=1.0\linewidth]{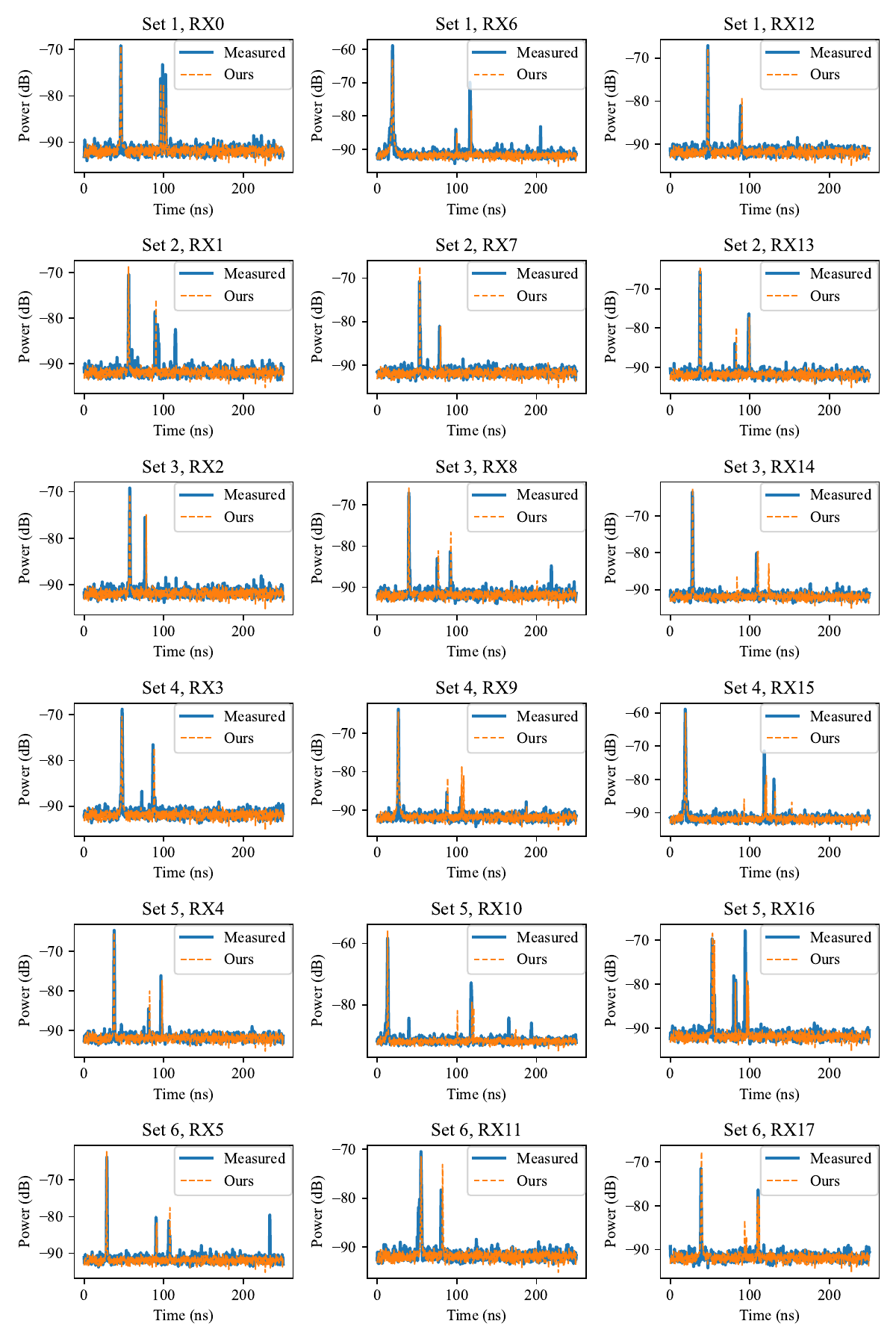}
    \caption{Qualitative results for all test RXs splits. The simulated ones are with applied noise computed from the last 50 samples of the measured RX0 PDP.}
    \label{fig:all_pdps}
\end{figure}

The RGB‑D data and the estimated camera poses exhibit substantial noise, as reflected by the average PSNR, SSIM, and LPIPS values of the training images: 20.73, 0.791, and 0.337, respectively. Qualitative results are provided in Fig.~\ref{fig:hall_render}. Despite these challenging conditions, our method successfully recovers a significant portion of the propagation paths present in the channel measurements, as illustrated in Fig.~\ref{fig:all_pdps}.

\paragraph{Ablations.}
We evaluate the contribution of each component in our loss $\mathcal{L}_{spa}$, namely the SD, the pruning of clustered taps, and the learnable gain coefficient $G_{ch}$ in Table~\ref{tab:ablation}. Without SD, the loss is forced to use the same tap index as in the measurements, even when it does not correspond to the true path. Without pruning, clustered weak taps may cause the optimizer to latch onto a non-specular MPC instead of the dominant peak. The learnable gain coefficient compensates for deviations from the isotropic assumption, as the aggregated measurements are not perfectly isotropic.

\begin{table}[H]
    \centering
    \caption{Ablations of components involved in $\mathcal{L}_{spa}$ and their impact of MAE of $\tau_{rms}$ in nanoseconds.}
    \begin{small}
    \begin{tabular}{lccccccc}
        \toprule
        Method & Set 1 & Set 2 & Set 3 & Set 4 & Set 5 & Set 6 & Mean \\
        \midrule
         Without SD & 14.10 & 9.34 & 9.19 & 8.60 & 12.89 & 16.73 & 11.81 \\
         Without prune & 11.78 & 7.42 & 5.88 & 7.88 & 10.82 & 12.81 & 9.43\\
         Without learnable $G_{ch}$ & 9.21 & 5.07 & 3.58 & 6.58 & 8.43 & 11.31 & 7.35 \\
         Full & \bf{7.97} & \bf{4.70} & \bf{3.20} & \bf{5.72} & \bf{6.85} & \bf{9.66} & \bf{6.35} \\
         \bottomrule
    \end{tabular}
    \end{small}
    \label{tab:ablation}
\end{table}

\section{Additional Experiments}
\label{appendix:additional_experiments}

In this appendix, we present additional experiments that further validate our approach, including novel view synthesis, synthetic radio‑propagation evaluation, and an investigation demonstrating that commercial smartphones provide sufficiently rich depth data to support reliable Gaussian‑based scene reconstruction for radio propagation simulation.

\subsection{Novel View Synthesis}
We evaluate the novel view synthesis quality of our method, as it imposes a limit of 128 Gaussians per ray and includes minor adjustments to the densification procedure. It also does not use the full Gaussian support, as we construct an approximation from 8 faces, as shown in Fig.~\ref{fig:overview}.

\begin{figure}[]
    \centering
    \includegraphics[width=1.0\linewidth]{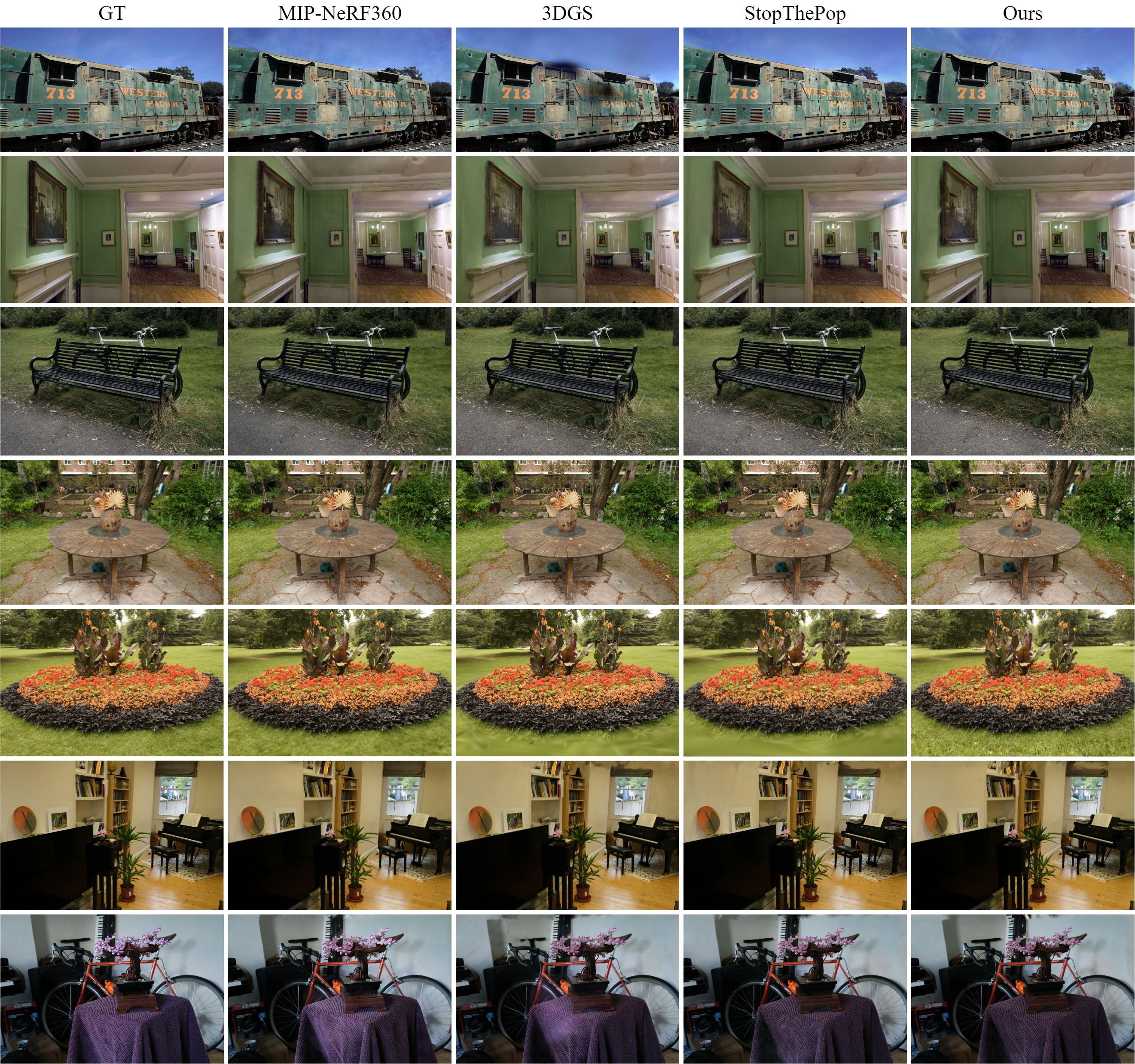}
    \caption{Novel views rendered with different methods. Mip-NeRF360 and 3DGS results were taken from \cite{kerbl20233d} and StopThePop from \cite{radl2024stopthepop}.}
    \label{fig:nvs}
\end{figure}

As the datasets for validating novel view synthesis, we use Mip-NeRF360 \cite{barron2021mip} dataset, Deep Blending \cite{hedman2018deep} and Tanks\&Temples \cite{knapitsch2017tanks} as in previous implicit and explicit NVS papers. These datasets contain a variety of different indoor and outdoor sceneries. For each dataset, we follow the standard train/test splits used in previous works. We compare against Mip-NerF360\cite{barron2021mip} 3DGS \cite{kerbl20233d}, 2DGS, \cite{huang20242d}, StopThePop \cite{radl2024stopthepop}, and 3DGRT \cite{moenne20243d}. For a fair comparison, we only optimize with $\mathcal{L}_{rgb}$. As shown in Table~\ref{tab:nvs_results}, our method achieves similar quality with the baselines, while using substantially fewer Gaussians, averaging under one million per scene. We also provide qualitative results in Fig.~\ref{fig:nvs}, which supports the claim of similar visual quality as the baseline methods. In terms of training time, our method offers a clear advantage over 3DGRT on MipNeRF360, reducing training duration considerably, as shown in Table~\ref{tab:training_time}. Both methods were trained with an NVIDIA GeForce RTX 5090 GPU.

\begin{table}[b]
    \centering
    \caption{Novel view synthesis metrics on Deep Blending, Mip-NeRF360, and Tanks\&Temples datasets. We compare both implicit and explicit methods for novel view synthesis with PSNR, SSIM and LPIPS metrics. All of the results were taken from the respective papers, except the 2DGS \cite{huang20242d} Tanks\&Temples and Deep Blending results were produced by us.}
    \resizebox{\linewidth}{!}{
    \begin{tabular}{cccccccccccccc}\toprule
         Dataset &  \multicolumn{3}{c@{}}{Outdoor (Mip-NeRF360)} & \multicolumn{3}{c@{}}{Indoor (Mip-NeRF360)} & \multicolumn{3}{c@{}}{Tanks\&Temples} & \multicolumn{3}{c@{}}{Deep Blending} & \\
         \cmidrule(lr){0-0}\cmidrule(lr){2-4}\cmidrule(lr){5-7}\cmidrule(lr){8-10}\cmidrule(lr){11-13}
         Metric  & PSNR~$\uparrow$ & SSIM~$\uparrow$ & LPIPS~$\downarrow$ & PSNR~$\uparrow$ & SSIM~$\uparrow$ & LPIPS~$\downarrow$ & PSNR~$\uparrow$ & SSIM~$\uparrow$ & LPIPS~$\downarrow$ & PSNR~$\uparrow$ & SSIM~$\uparrow$ & LPIPS~$\downarrow$ &Avg. \#Gaussians \\ 
         \midrule
          Mip-NeRF360 &  24.39 & 0.691 & 0.287 & 31.58 & 0.914 &  0.182 & 22.22 & 0.759 & 0.257 & 29.40 & 0.901 & 0.245 & -\\
         3DGS & 24.64 &   0.731 &  0.234 & 30.41 &   0.920 &  0.189 &  23.14 & 0.841 &  0.183 & 29.41 &  0.903 &  0.243 & 3.06~M\\
         2DGS & 24.34 & 0.717 &  0.246 & 30.40 & 0.916 & 0.195 & 23.13 &  0.832 & 0.213 & 29.57 &  0.904 & 0.257 &  1.75~M\\
         StopThePop & 24.46 &  0.722 & 0.254 & 30.03 & 0.917 & 0.194 &  23.18 &  0.839 &  0.184 &  29.84 &  0.905
         &  0.241 &  1.54~M \\
         3DGRT & 24.34 &  0.721 & - & 30.23 & 0.920 & - & 23.20 &  0.830 & 0.222 &  29.23 & 0.900 & 0.315 &  -\\
         Ours & 24.11 &  0.721 & 0.241 & 29.55 &  0.911 & 0.195 & 22.60 &  0.837 & 0.193 &  30.03 & 0.908 & 0.245 &  0.96~M\\
         \bottomrule
    \end{tabular}
    }
    \label{tab:nvs_results}
\end{table}

\begin{table}[]
    \centering
    \caption{Average training times in Mip-NeRF360, Tanks\&Temples and Deep Blending scenes.}
    \begin{small}
    \begin{tabular}{lcccc}\toprule
         Method & Mip-NeRF360 & Tanks\&Temples & Deep Blending\\
         \midrule
         3DGRT & 47.8~min & - & -\\
         Ours & 27.2~min & 13.3~min & 23.7~min\\
         \bottomrule
    \end{tabular}
    \end{small}
    \label{tab:training_time}
\end{table}

\subsection{Radio Propagation Validation with Synthetic Data}

\begin{figure}[]
    \centering
    \begin{subfigure}{0.49\linewidth}
        \centering
        \includegraphics[width=\linewidth]{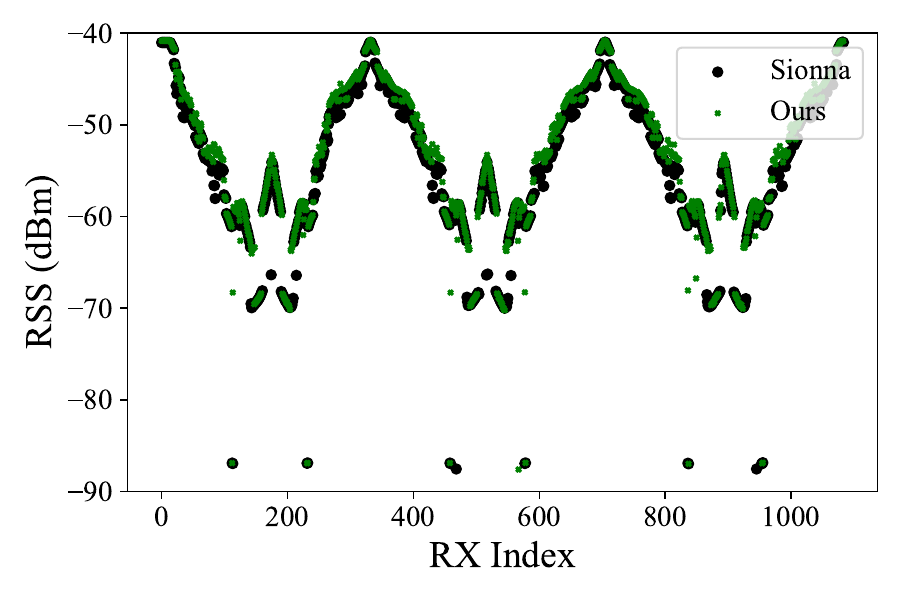}
        \caption{Received signal strength from the line of sight and first-order reflection paths.}
        \label{fig:syn_1st_order}
    \end{subfigure}
    \hfill
    \begin{subfigure}{0.49\linewidth}
        \centering
        \includegraphics[width=\linewidth]{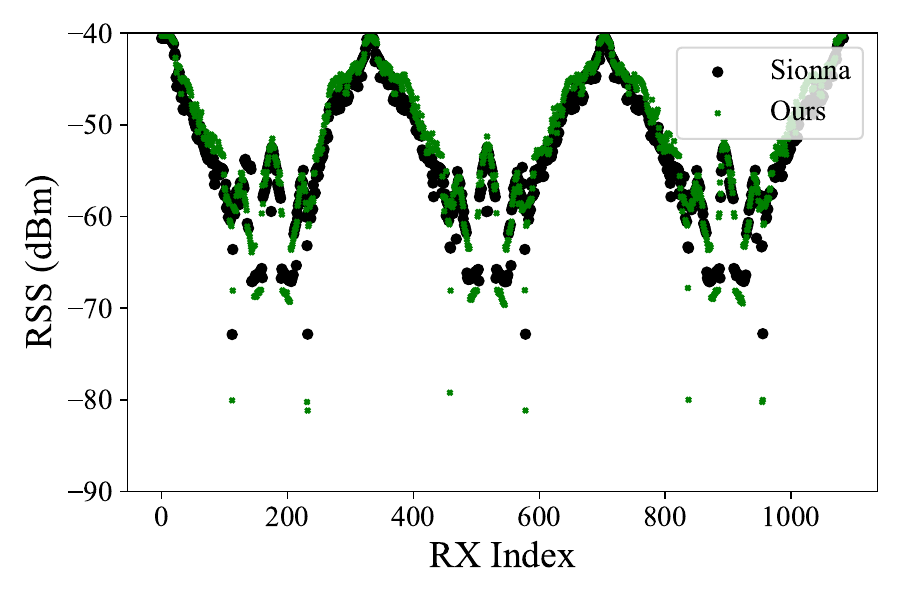}
        \caption{Received signal strength from the line of sight and up to second-order reflection paths.}
        \label{fig:syn_2nd_order}
    \end{subfigure}

    \caption{Received signal strength predictions at 1084 receiver locations provided in the RF3DGS \cite{zhang2026rf} dataset with our method and with Sionna.}
    \label{fig:syn_rss}
\end{figure}

To further assess the fidelity of our Gaussian‑based radio propagation simulator and its ability to generalize to complex synthetic environments, we design a controlled evaluation that compares its predictions against those of Sionna \cite{hoydis2023sionna} under identical geometric and propagation conditions. This experiment allows us to validate whether a learned Gaussian scene representation can accurately reproduce physically grounded multipath behavior.
we utilize triangle mesh models provided in RF3DGS \cite{zhang2026rf} dataset to compare our simulation results with Sionna. Using these meshes, we generate segmentation, normal, and depth images in addition to the RGB images included in the dataset and train our Gaussian model with these. For evaluation, we perform RT at each of the 1084 RX coordinates provided in the dataset using both our method and Sionna \cite{hoydis2023sionna}. From the ray traced paths we compute received signal strength (RSS) values as non-coherent sum with
\begin{equation}
\label{eq:rss}
    RSS = 30+ 10 \log_{10}\left(P_{tx} \sum_{i=1}^N |a_{i}|^2\right).
\end{equation}
We evaluate two simulation cases, both including LoS paths: one allowing up to first-order reflections and another allowing up to second-order reflections. Qualitative and quantiative results are shown in Fig.~\ref{fig:syn_rss}, and Table~\ref{tab:syn_rss}, demonstrating agreement between our RSS predictions and those obtained with Sionna.

\begin{table}[]
    \centering
    \caption{Comparison of mean and standard deviation of received signal strength at 1084 receiver locations.}
    \begin{small}
    \begin{tabular}{ccccc}
        \toprule
        & \multicolumn{2}{c}{1 reflection \& LoS} 
        & \multicolumn{2}{c}{2 reflections \& LoS} \\
        \cmidrule(lr){2-3}\cmidrule(lr){4-5}
        & Mean (dBm) & Std (dBm) & Mean (dBm) & Std (dBm) \\
        \midrule
        Ours   & -53.43 & 8.22 & -52.31 & 8.05 \\
        Sionna & -54.64 & 8.71 & -53.14 & 7.41 \\
        \bottomrule
    \end{tabular}
    \end{small}
    \label{tab:syn_rss}
\end{table}

\begin{figure}[H]
    \centering
    \includegraphics[width=1.0\linewidth]{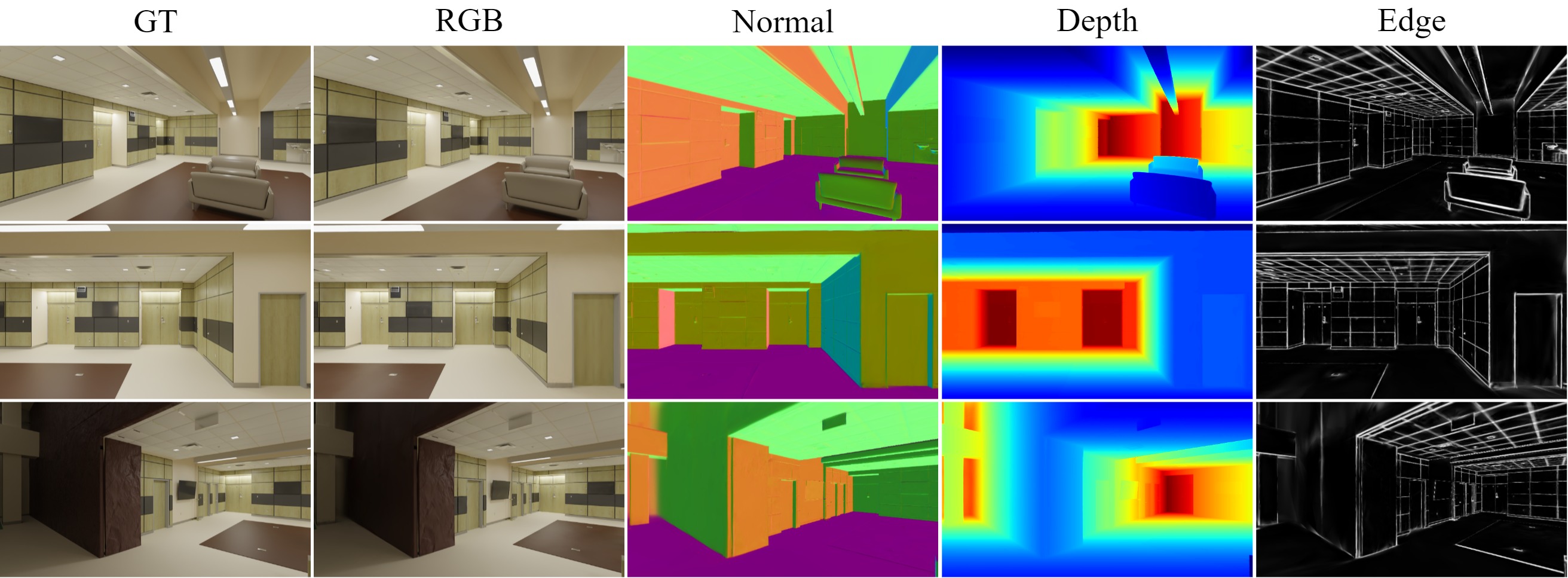}
    \caption{Ground truth RGB image, as well as the rendered RGB, normal, depth and edges by our method in the RF3DGS scene. Zoom in for details.}
    \label{fig:rf_render}
\end{figure}

\subsection{Real-World RSS Experiments}

To enable practical and accessible scene reconstruction for wireless propagation modeling, we leverage data captured directly from a modern smartphone, demonstrating that such devices can provide sufficiently rich visual and depth information for building Gaussian‑based scene representations. We then utilize the reconstructed scene to perform RT and showcase how RSS measurements can be used to learn the channel parameters.

\paragraph{Scene Description.}

\begin{figure}[H]
    \centering

    \begin{subfigure}{0.49\linewidth}
        \centering
        \includegraphics[width=\linewidth]{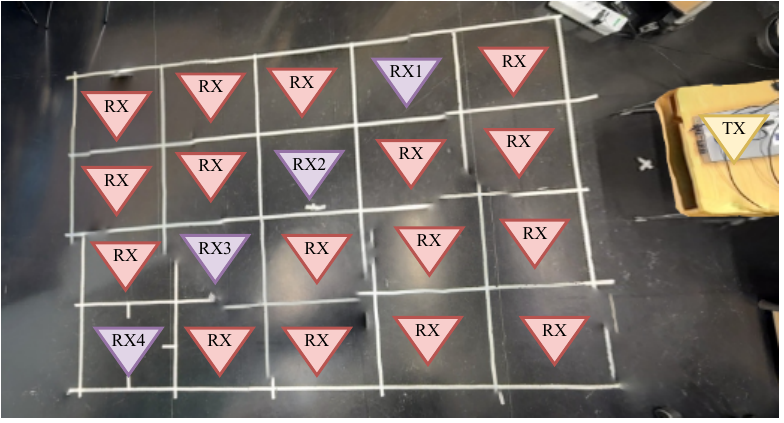}
        \caption{Measurement locations in the laboratory. The purple RXs (RX1-RX4) are used for validation.}
        \label{fig:mms_lab}
    \end{subfigure}
    \hfill
    \begin{subfigure}{0.41\linewidth}
        \centering
        \includegraphics[width=\linewidth]{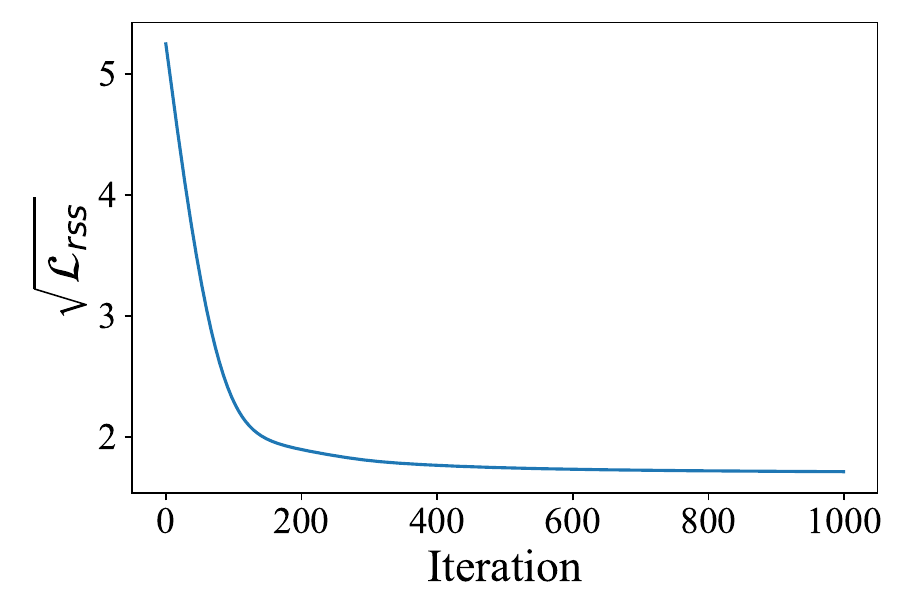}
        \caption{Per iteration loss graph for the laboratory experiments for the Red, unnumbered RXs.}
        \label{fig:mms_loss}
    \end{subfigure}

    \caption{Measurement setup in the scene and loss graph for the experiments in this environment.}
    \label{fig:mms_res}
\end{figure}

The RSS measurements were collected at each cell in Fig.~\ref{fig:mms_lab} using two USRP N310 devices operating at a center frequency of 3.85~GHz with a 60~MHz bandwidth. Both TX and RX used dipole antennas.

Camera data was captured with an iPhone, which uses the iPhone depth information to generate camera poses in metric scale. From our observation the camera poses were too inaccurate for Gaussian-based reconstruction, which is why we resorted to utilizing COLMAP \cite{schoenberger2016mvs, schoenberger2016sfm}. As COLMAP only uses color information, the metric scale has to be resolved separately due to scale ambiguity. Following Umeyama’s least‑squares similarity alignment method \cite{umeyama91least}, we estimate the global rotation, translation, and scale that aligns the COLMAP camera locations to those obtained from iPhone. The semantic labels are applied to the Gaussians as described in Appendix~\ref{appendix:label}.

\paragraph{Training.}

We train the relative permittivity $\epsilon_{r}$ and conductivity $\sigma_{c}$ of each material and treat the transmitted power $P_{tx}$ as an additional optimizable parameter. The number of reflections is limited to two and the RSS is computed with Eq.~(\ref{eq:rss}).
All material parameters are initialized to plasterboard at 3.85~GHz based on ITU-R-P.2040 \cite{itu}. The transmitted power $P_{tx}$ is initialized to 5~dBm.
As our loss function we use the mean squared error (MSE) loss
\begin{equation*}
    \mathcal{L}_{rss} = \frac{1}{N} \sum_{i=1}^N (RSS_{i} - \hat{RSS}_{i})^2,
\end{equation*}
where $RSS$ and $\hat{RSS}$ are theground truth and predicted RSS values. We run the training for 1000 iterations using the red RXs shown in Fig.~\ref{fig:mms_lab}.

\begin{figure}[]
    \centering
    \includegraphics[width=0.8\linewidth]{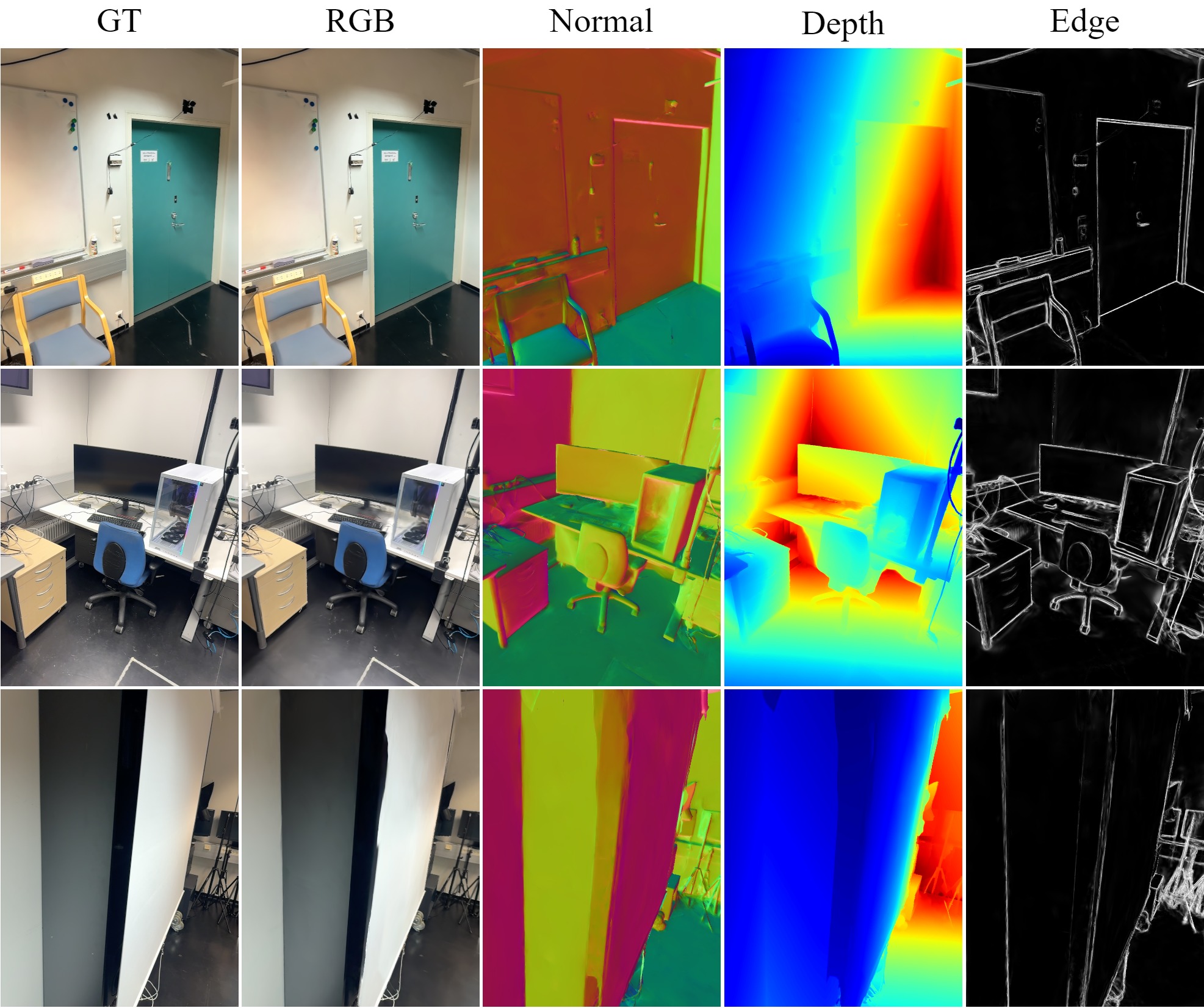}
    \caption{Ground truth RGB image, as well as the rendered RGB, normal, depth and edges by our method in the room scene. Zoom in for details.}
    \label{fig:mms_render}
\end{figure}

\paragraph{Results.}
We evaluate performance on the test set (yellow RXs in Fig.~\ref{fig:mms_lab}). As shown in Table~\ref{tab:mms_res}, our method achieves a mean absolute error of roughly 0.4\,dBm, closely matching the measured RSS values. The loss graph in Fig.~\ref{fig:mms_loss} further show stable and well‑behaved convergence in this scenario. We also provide some qualitative results in Fig.~\ref{fig:mms_render}.

\begin{table}[H]
    \centering
       \caption{Predicted RSS values for the test set before and after optimization}
    \begin{small}
    \begin{tabular}{ccccc}
        \toprule
          RSS (dBm) & RX 1 & RX 2 & RX 3 & RX 4 \\
         \midrule
         Measured & -35.7 & -38.4 & -42.1 & -43.5 \\
         Initial & -41.4 & -43.8 & -46.4 & -48.6 \\
         \midrule
         Optimized & -36.2 & -38.6 & -41.4 & -43.7\\
         Absolute Error & 0.49 & 0.23 & 0.67 & 0.25\\
         \bottomrule
    \end{tabular}
    \end{small}
    \label{tab:mms_res}
\end{table}

\subsection{Assessment of Edge-Aware Ray Tracing}
\label{appendix:edge_aware}
\begin{figure}[]
    \centering
    \includegraphics[width=0.9\linewidth]{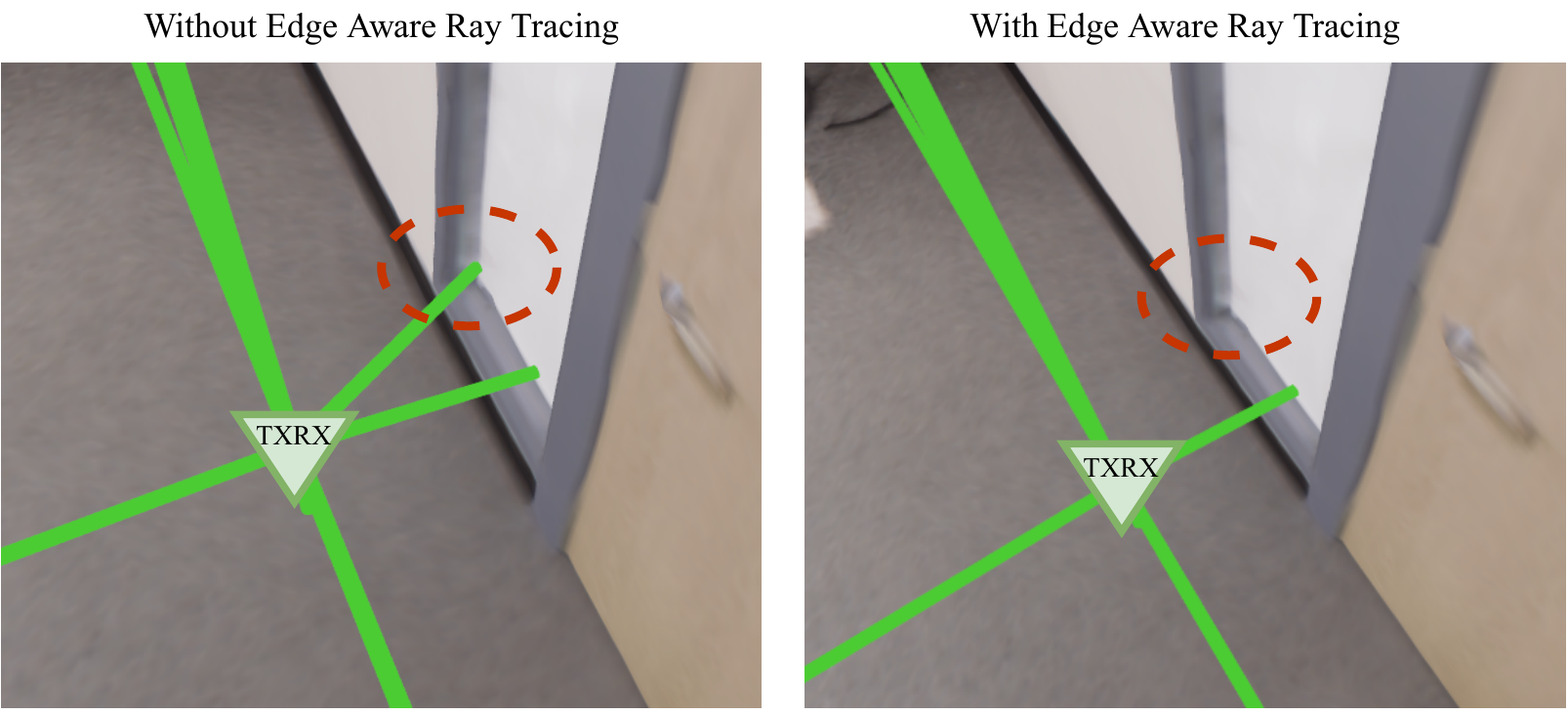}
    \caption{Ray traced paths with and without edge-aware ray tracing.}
    \label{fig:edge_aware}
\end{figure}

We highlight the importance of our edge-aware ray tracing by training a high-quality model on \textit{room0} from the Replica dataset~\cite{replica19arxiv}, which provides ground-truth depth and normal maps. In Fig.~\ref{fig:edge_aware}, we place the TX and RX at the same location and simulate only first-order reflections. Even in this ideal synthetic setting, paths can incorrectly converge to geometric edges, producing non-physical specular reflections. Our method addresses this issue by leveraging monocular normal estimates to learn per-Gaussian edge probabilities through the loss term $\mathcal{L}_{e}$. During point-to-point path tracing, we combine these learned edge probabilities $\bar{\gamma}$ with normal coherency $\|\bar{n}\|$ to prune invalid paths. A path is discarded if the weighted average edge probability of its intersected Gaussians exceeds the threshold $\bar{\gamma} > g_{\bar{\gamma}}$, or if the aggregated normal vector contains insufficient coherency, i.e., $\|\bar{n}\| < g_{c}$. This edge-aware pruning effectively removes these physically invalid specular reflections.

\end{document}